# High-Fidelity Longitudinal Patient Simulation Using Real-World Data

Yu Akagi[1*], Tomohisa Seki[2], Hiromasa Ito[2,3], Toru Takiguchi[2,4], Kazuhiko Ohe[5,6], Yoshimasa Kawazoe[2,5*]

[1] Department of Biomedical Informatics, Graduate School of Medicine, The University of Tokyo, Tokyo, Japan
[2] Department of Healthcare Information Management, The University of Tokyo Hospital, Tokyo, Japan
[3] Department of Cardiology and Nephrology, Mie University Graduate School of Medicine, Mie, Japan
[4] Department of Emergency and Critical Care Medicine, Nippon Medical School, Tokyo, Japan
[5] Artificial Intelligence and Digital Twin in Healthcare, Graduate School of Medicine, The University of Tokyo, Tokyo, Japan
[6] Graduate School of Health Data Science, Juntendo University, Tokyo, Japan

## Abstract
Simulation is a powerful tool for exploring uncertainty. Its potential in clinical medicine is transformative and includes personalized treatment planning and virtual clinical trials. However, simulating patient trajectories is challenging because of complex biological and sociocultural influences. Here, we show that real-world clinical records can be leveraged to empirically model patient timelines. We developed a generative simulator model that takes a patient's history as input and synthesizes fine-grained, realistic future trajectories. The model was pretrained on more than 200 million clinical records. It produced high-fidelity future timelines, closely matching event occurrence rates, laboratory test results and temporal dynamics in real patient future data. It also accurately estimated future event probabilities, with observed-to-expected ratios consistently near 1.0 across diverse outcomes and time horizons. Our results reveal the untapped value of real-world data in electronic health records and introduce a scalable framework for in silico modelling of clinical care.

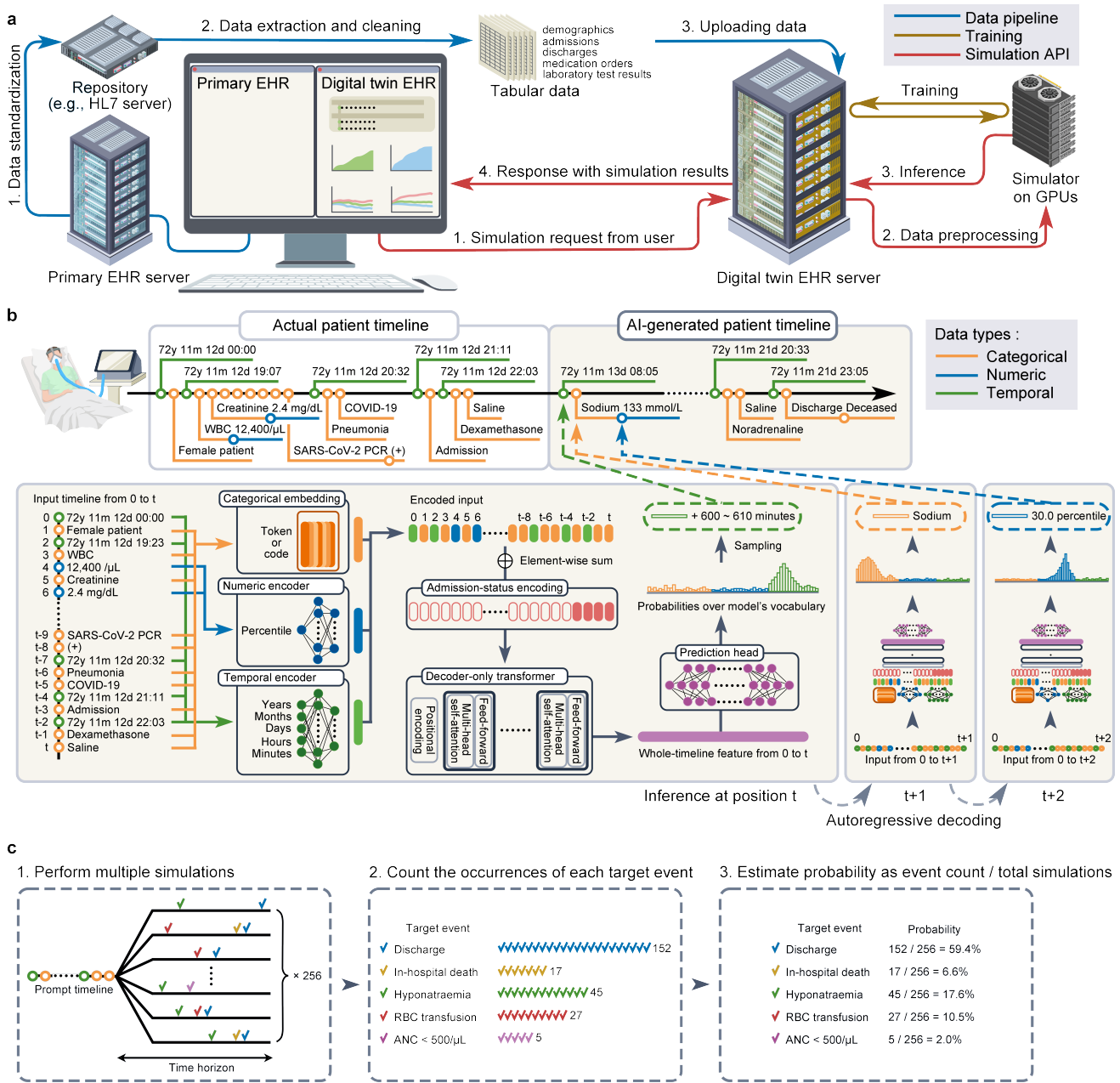

**Fig. 1 | Framework for clinical trajectory simulation using real-world data. a**, Overview of the simulation framework. RWD are extracted from EHR servers using a standardized format and uploaded to the digital twin server. The simulator model is trained using these data. The framework is designed so that it can be integrated into clinical workflows via API (red arrows). **b**, Patient timeline and model architecture. We structured clinical records into a timeline and defined each point in the timeline as an 'entry'. A patient timeline is input into the simulator, and future timelines are generated autoregressively. Distinct encoders are used for categorical, numeric, and temporal features. The encoded inputs are combined with admission-status encodings and passed through the decoder-only transformer stack. The model predicts probabilities over its vocabulary and samples one token per step. The detailed structure of a timeline is illustrated in Extended Data Fig. 1. **c**, Event prediction using the Monte Carlo method. For each prompt, the model generates multiple timelines. Event probabilities are estimated as the occurrence rate across simulations. AI, artificial intelligence; API, application programming interface; EHR, electronic health record; GPU, graphics processing unit; HL7, Health Level Seven; RWD, real-world data; WBC, white blood cell count.

# Introduction

Simulation is a powerful tool for exploring possibilities. For example, model training through driving simulation enabled the deployment of self-driving taxis in San Francisco[1]. Protein folding simulation using AlphaFold advanced scientific research, recognized with a Nobel Prize[2]. Simulation is also expected to transform clinical medicine. If accurate patient simulation becomes possible, clinical trials could be conducted in silico[3–5]. Clinical simulation may enable counterfactual predictions[6]. Additionally, simulation could play a vital role in personalized medicine by predicting individual disease risks and treatment effects[3,7,8].

However, patient simulation is currently limited by the complexity of clinical medicine. A clinical course is inherently a longitudinal trajectory that reflects biological interactions within the human body[4]. Patient care also involves sociocultural systems. Therefore, it is challenging to theoretically simulate the entire system. However, it may be possible to model patient timelines empirically by leveraging clinical records from real-world practice. Such data are referred to real-world data (RWD). Although RWD have long existed, their use in patient simulation have been limited. One reason is that RWD are typically siloed within local electronic health record (EHR) systems with inconsistent formats, thereby limiting scalable data collection[9]. Another obstacle is the lack of a method to generate complex time series data such as patient trajectories. A patient's clinical course consists of various types of records recorded at uneven time intervals[10]. Overcoming these limitations requires a new technical framework. Recent advances in artificial intelligence (AI) methods and data infrastructure offer promising solutions to these challenges.

RWD are becoming increasingly available at scale. In the United States, medical data sharing standards such as Health Level Seven (HL7) have been updated and widely implemented. These standards have been integrated into major analytics platforms such as i2b2 (Informatics for Integrating Biology and the Bedside)[11,12]. Japan also has a medical data standard built upon HL7. This standard is implemented in more than 1,000 Japanese hospitals[13]. The European Commission is developing the European Health Data Space, a large virtual data space within which to share RWD across Europe[14]. Such international efforts are making RWD more scalable and emphasizing the need for robust frameworks to utilize large amounts of clinical data meaningfully.

Longitudinal patient simulation using RWD can be realized with emerging generative AI techniques. A promising approach is autoregressive decoding[15,16]. This method is the core mechanism of large-language models (LLMs). These models produce an object in the text (or 'token') repeatedly. The generated object is appended to the existing text, and the updated text is used as input for the next prediction. The model repeats these steps until an end condition is met. This mechanism is enabled by a modern AI architecture known as a 'decoder-only transformer'[16].

This algorithm can be adapted for longitudinal clinical simulation. A plausible idea is to replace text tokens with clinical timeline entries. By sequentially generating patient medical records as a timeline, it may be possible to create a simulated patient trajectory. Studies exploring similar ideas have shown potential[17,18]. For example, a model trained based on medical concepts extracted from clinical notes generated a chronological series of concepts deemed clinically relevant by physicians[17]. Another study revealed that autoregressive modelling of medical events could aid in predicting in-hospital deaths[18]. However, these early attempts struggled to generate time series data with practical granularity. In clinical practice, timestamps are recorded to the level of minutes, and laboratory test results include numeric values. The transformer decoder, originally designed for discrete variables, encounters inherent challenges in generating data with continuous variables. Furthermore, how to format clinical records into a single timeline is an unresolved question. A new, effective framework is needed.

In this study, we propose a framework for longitudinal patient simulation using RWD (Fig. 1). The framework incorporates a generative model that produces fine-grained clinical trajectories, including minute-level timestamps and precise numeric laboratory values and units. We trained and evaluated the model using data from 356,062 patients in a real-world clinical setting. The simulator generated high-fidelity future trajectories and showed strong calibration across diverse clinical events and time horizons. These results support the feasibility of empirical longitudinal patient simulation using large-scale RWD.

## Results
### Data acquisition
We extracted real-world clinical records from EHR servers. Supplementary Fig. 1 summarizes the dataset development and patient selection processes. Patient and timeline characteristics are summarized in Extended Data Table 1. We collected clinical data from January 2011 to December 2023 via standardized clinical data storage at the University of Tokyo Hospital using HL7 version 2.5 messages. We obtained 292,210,334 records, 1.85% of which (5,408,923) were excluded during data cleaning. A total of 373,890 patients had at least one valid clinical record during this period. To create the test dataset, we first reserved the newest 25,735 (7%) patients on the basis of the first date in their records. The remaining patients were then randomly split into training (n=340,659) and validation (n=7,496) datasets.

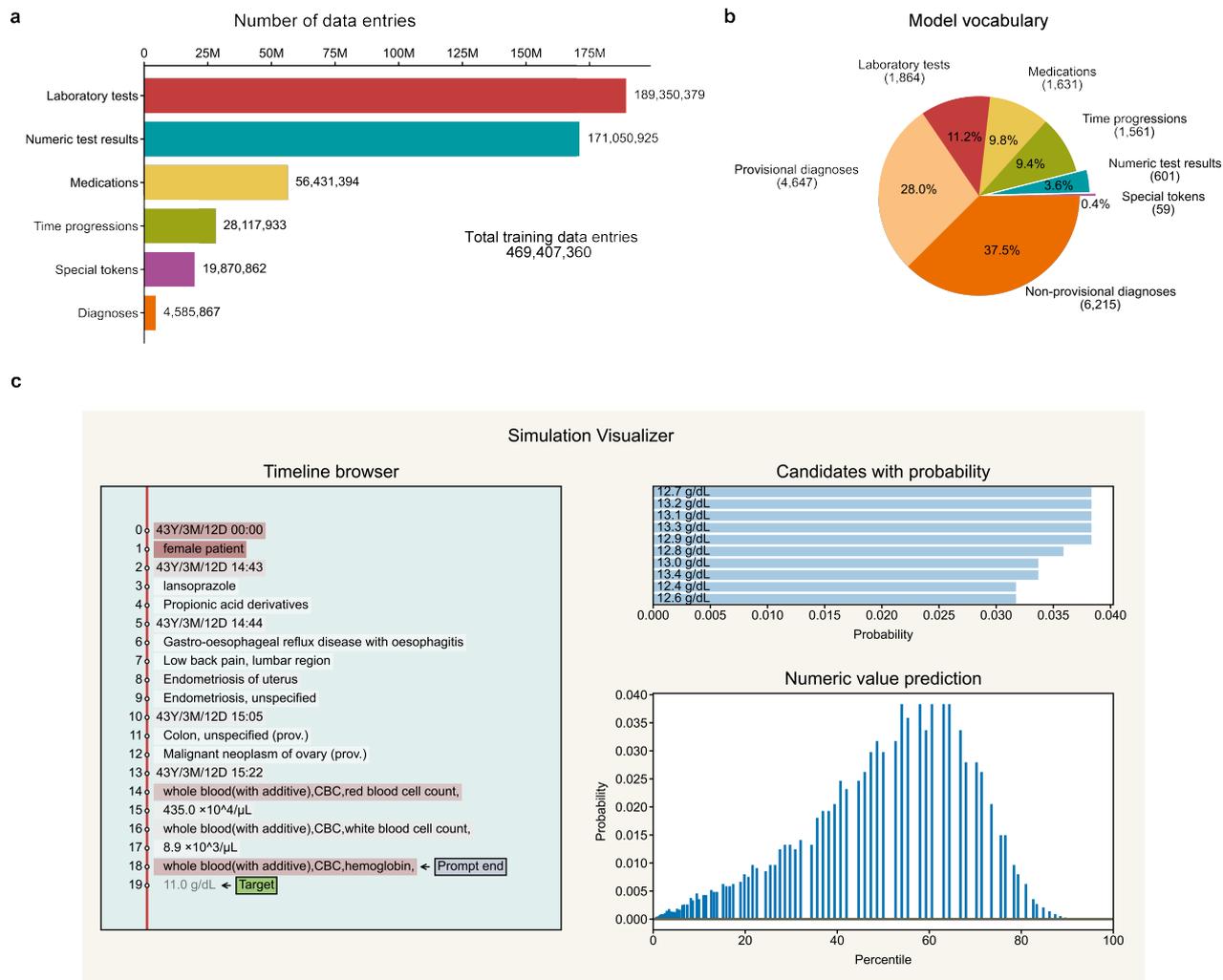

**Fig. 2 | Summary of model development results. a**, Distribution of timeline entries used for training. The model was trained using a dataset of 469,407,360 timeline entries. **b**, Composition of the model's vocabulary. The vocabulary includes 16,578 unique entries. The list of special tokens appears in Supplementary Table 1. **c**, Visualization of an inference step. The left panel shows the simulated timeline in the timeline browser. Prompt entries (i.e., model inputs) are shown in dark text. Red shading indicates tokens that received high attention weights. Future entries are shown in grey. The top-right panel shows predicted top candidates from the model's vocabulary. The bottom-right panel shows the predicted distribution of a numeric test result. In this example, the model is predicting a haemoglobin value for a 43-year-old female patient. The model places high attention weights on her age and sex, leading to a predicted haemoglobin distribution skewed towards lower values. This approach aligns with clinical knowledge that premenopausal women often exhibit lower haemoglobin levels due to menstrual blood loss[63]. Additional examples are shown in Supplementary Fig. 2.

## Model development

We developed a model that simulates future clinical trajectories conditioned on a patient's prior timeline (Fig. 1b). This model uses a transformer architecture and follows an autoregressive decoding approach. Its unique aspect is that it has dedicated encoders for temporal and numeric inputs. The model also has a unique ability to decode numeric laboratory test results with appropriate decimal accuracy and units and time at minute resolution. This unique architecture enables the representation and decoding of clinical timelines at full clinical resolution. See Methods for details.

We trained the model using clinical records from January 2011 through December 2022. The model development summary is shown in Fig. 2. The initial pretraining used all the training records. The training was finished at epoch 82, with early stopping applied. We then fine-tuned the pretrained model using records from only 2022 to fit the model to recent clinical practice. The training concluded at epoch 12, with early stopping applied. The final vocabulary size, representing all unique entries the model could generate, reached 16,583. The model learned to generate patient timelines with fine resolution, as illustrated in Extended Data Fig. 2.

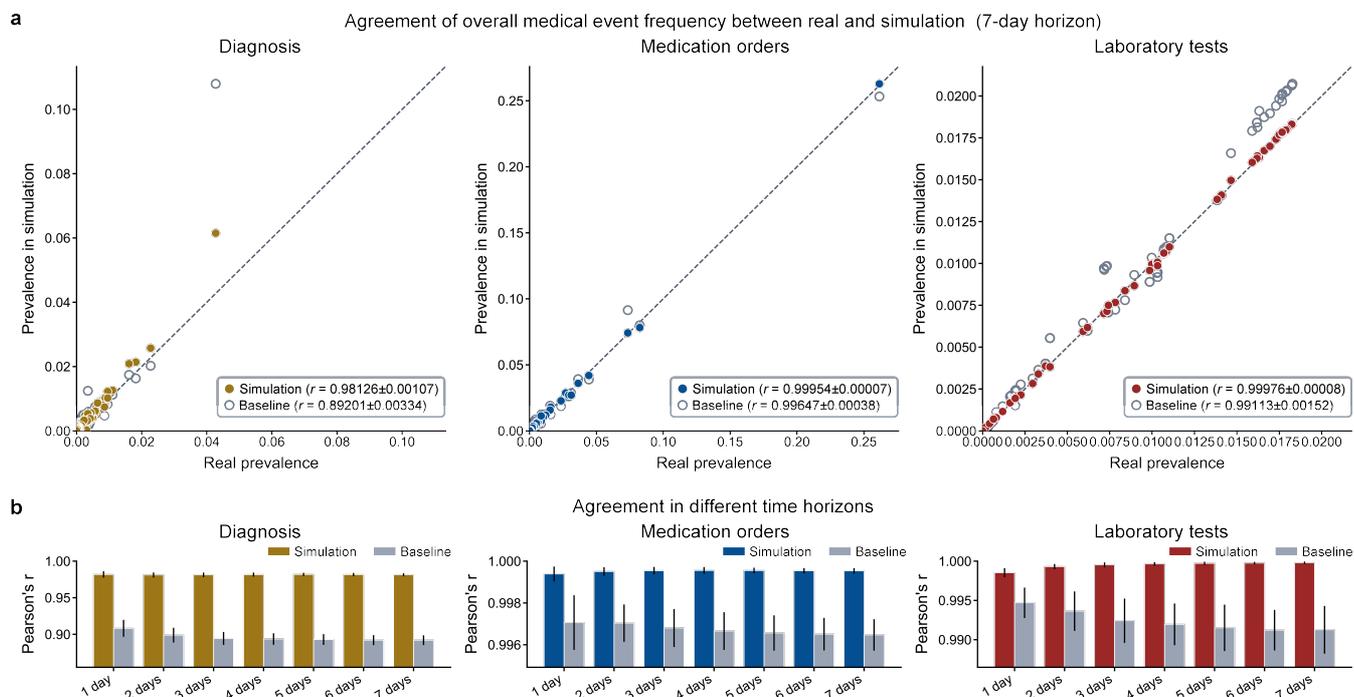

**Fig. 3 | The simulator accurately reproduced major medical event rates.** The figure illustrates the agreement of the prevalence of major medical events in reality and in the simulation, measured by Pearson's correlation coefficient ($r$). The model outperformed the baseline across different time horizons and event categories (diagnosis, medication orders, and laboratory test events), with high agreement with real data ($r >$ 0.98). The major events are defined as the most frequent medical events that accounted for 90% of the total events among in each category: 1,334 diagnosis codes, 293 medication codes, and 123 laboratory tests. **a**, Agreement of overall frequency of medical events in the 7-day time horizon (left, diagnoses; middle, medication orders; right, laboratory tests). Each dot represents a distinct event identified by a unique medical code. The frequency was defined as the total count of a code divided by the total number of codes in its category. Agreement is quantified by $r$ (reported with 95% confidence intervals). **b**, Agreement of overall frequencies in reality versus the simulation across different time horizons.

## Sampling diverse patient timelines and generating simulated timelines

We sampled diverse patient timelines to assess the model's simulation performance. We included 7,907 test patients who admitted in 2023. There were 12,606 admissions (Extended Data Table 1). For each patient, we collected pairs of a prompt timeline and its future timeline during hospitalization daily, up to day 30. On each day, a prompt-future pair was collected at 12:00 PM, when most routine laboratory test results are available. Timelines before this time were collected as prompts, and the timelines beyond this point as real future timelines. This sampling strategy covers diverse clinical states, ranging from acute illness at admission, to recovery before discharge, to critically ill near death (see Methods). In total, we obtained 103,579 prompt-future pairs.

Because patient timelines are not deterministic in nature, a single simulated timeline does not capture the entire possibilities that a prompt timeline can reveal in the future. Therefore, we ran 256 stochastic simulations per prompt. Each simulation continued for a 7-day time horizon. In total, we generated 26,516,224 simulated timelines. We excluded two timelines that did not complete even after 7,000 autoregressive steps to avoid excessive runtime. Timeline generation was completed in 228 hours and 25 minutes. This corresponds to a mean runtime of 104 seconds per patient and 65 seconds per admission. The mean runtime to generate 256 simulations per prompt was 8 seconds. We used these timelines for the downstream analysis. We also analyzed these timelines by different time horizons, from 1-day to 7-day horizon as sensitivity analyses.

## Fidelity of the simulated future timelines

We assessed the fidelity of the generated timelines by comparing them to real future timelines (n = 103,579). For each prompt, one simulated timeline was randomly sampled from the 256 generated candidates. This process was repeated 256 times, yielding 256 simulated timeline sets, each matched in size to the real set.

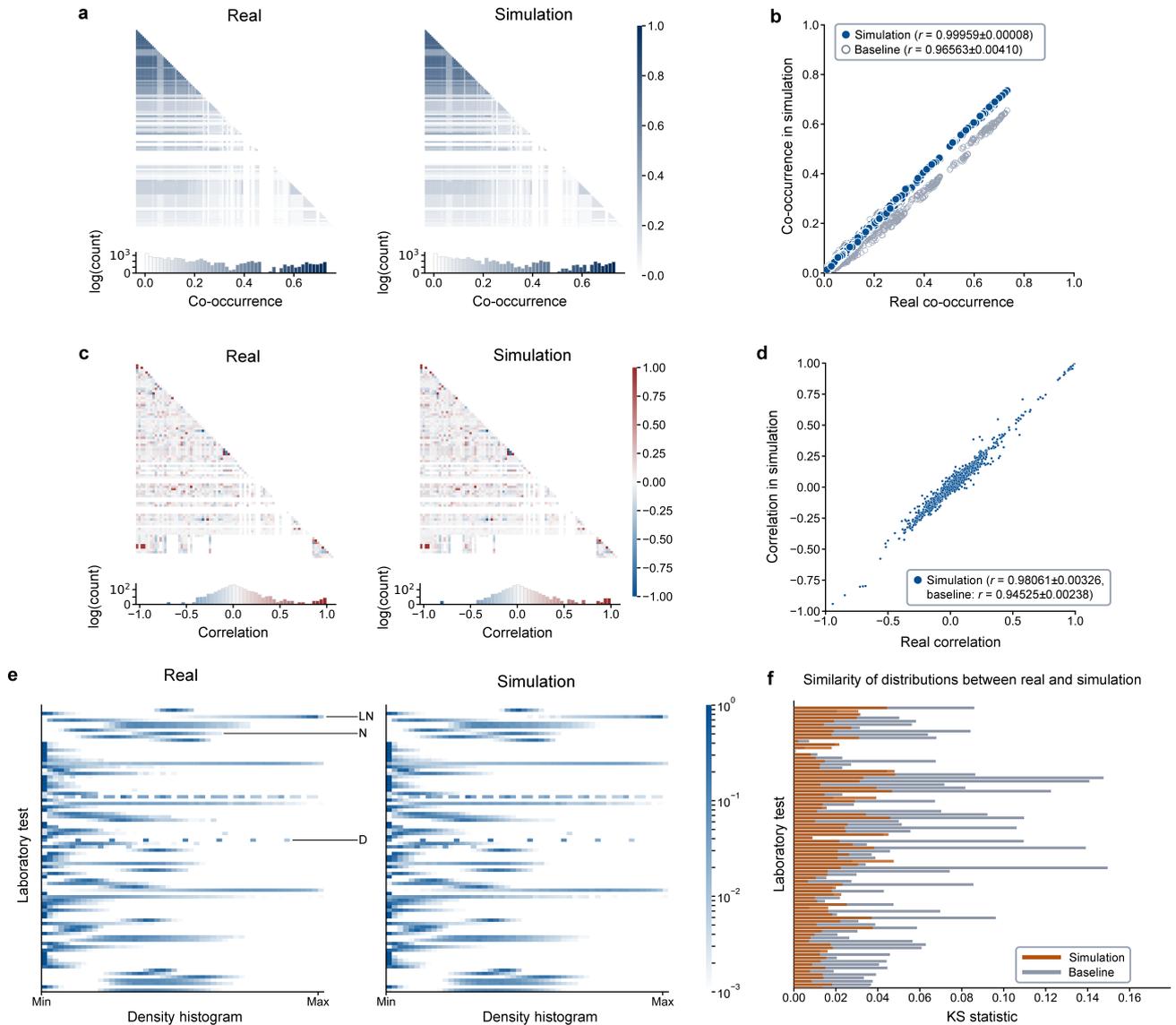

**Fig. 4 | The simulator can generate high-fidelity and diverse laboratory test data over patients' future clinical courses. a**, Co-occurrence matrix of 123 major laboratory tests, capturing test pair frequencies across simulated timelines. **b**, Agreement of co-occurrence frequencies between real and simulated data ($n = 7{,}503$). **c**, Correlation matrix of numeric test results from 85 major laboratory tests. Pairs with insufficient observations for Pearson's correlation are shown as blank. **d**, Agreement between real and simulated intertest correlations ($n = 2{,}212$). **e**, Distribution of numeric test values for the 85 major laboratory tests. Each row shows a normalized density histogram for one test. Representative distribution types are highlighted: normal-like ('N'), log-normal-like ('LN'), and discrete-like ('D'). **f**, Kolmogorov–Smirnov (KS) statistics quantifying the similarity between real and simulated distributions shown in **e**. The y-axis is shared with **e**. The median KS statistic was lower in the simulation than the baseline (0.01785 [95% CI: 0.01529–0.02626] vs. 0.03972 [95% CI: 0.03752–0.04156]). Pearson's correlation coefficients ($r$) are reported with 95% confidence intervals.

Fidelity metrics were computed for each simulated set and summarized using 95% confidence intervals (see Methods).

As a reference, we constructed a noncontextual baseline model that exactly reproduces the empirical distribution of the 2022 training data, to which the generative model was fine-tuned. This baseline represents an upper limit for noncontextual fidelity and is commonly used in EHR synthesis studies[19]. Performance exceeding this baseline indicates that the model captures context-dependent patterns rather than simply reproducing memorized distributions.

The simulated timelines demonstrated high fidelity in reproducing the occurrence rates of major medical events (Fig. 3). We first evaluated the overall frequency of medical events, defined as the proportion of each distinct event among all events within each group (diagnoses, medication orders and laboratory tests). The simulated prevalence closely matched real-world distributions across all groups (Fig. 3). The agreement between the simulated and real occurrence rates, measured by Pearson's correlation coefficient (r), was consistently high (r > 0.981) and exceeded that of the baseline model across all groups and time horizons. We also assessed the event rates at the individual timeline level (Supplementary Fig. 4). We defined this timeline-level occurrence rate as the probability that a distinct event appeared at least once within a timeline. The simulator outperformed the baseline and strongly agreed with the real data across all event types and time horizons.

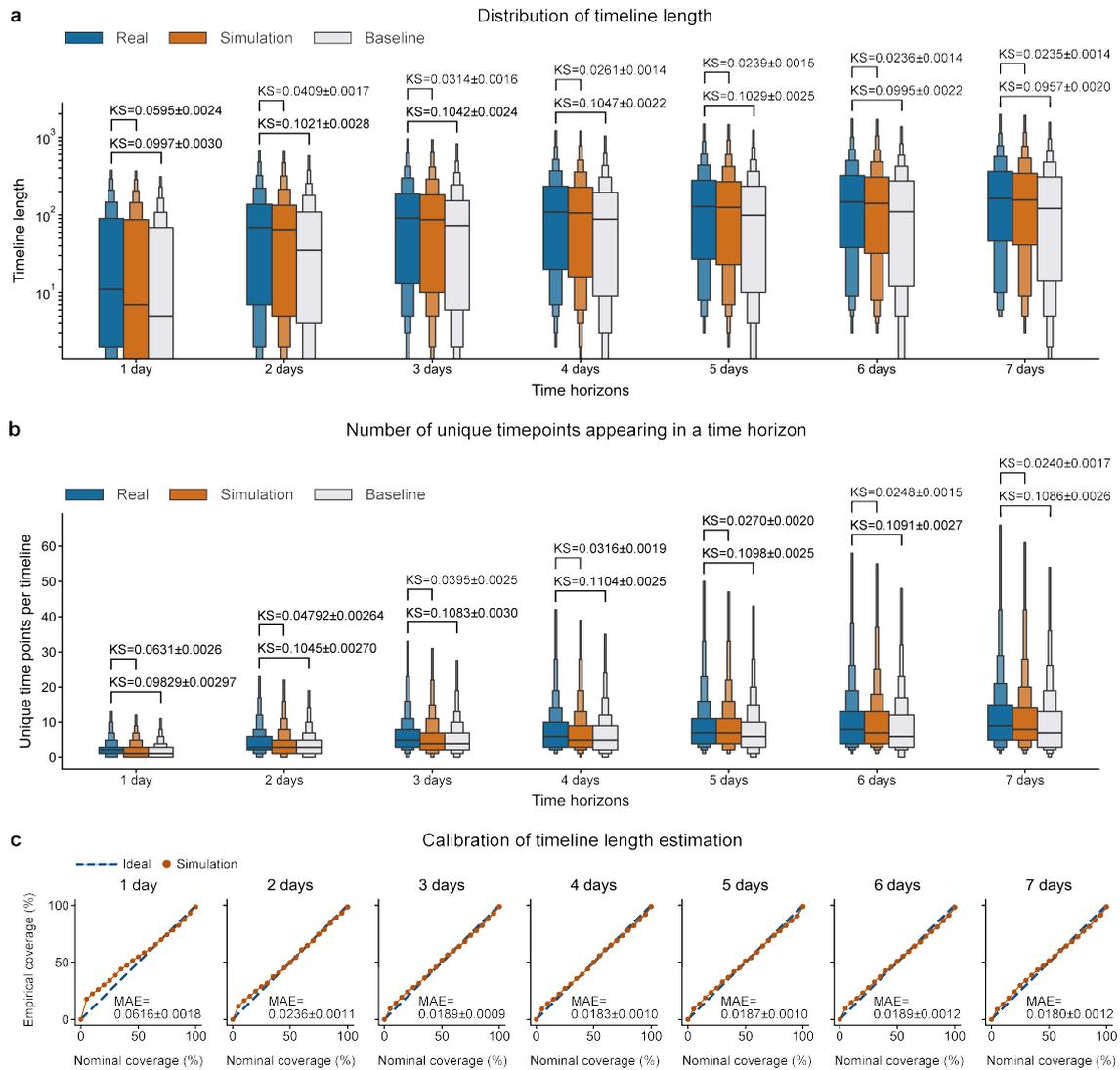

**Fig. 5 | The simulator captured the longitudinal dynamics of patient timelines.** The model accurately reproduced the distribution of timeline lengths and temporal entries. It demonstrated well-calibrated uncertainty in simulated timeline durations. **a**, Timeline length distributions in real and simulated data. The box plots show the median and interquartile range (central box), with extended boxes representing the 12.5–87.5%, 6.25–93.75%, and 3.125–96.875% percentile ranges. The distributions were well reproduced in simulation with small KS statistics. The simulator consistently outperformed the baseline. **b**, Distribution of the number of temporal entries per timeline in reality and in the simulation. The simulator demonstrated small KS statistics and outperformed the baseline across all time horizons. **c**, Coverage probability analysis for timeline length. The close alignment between nominal and empirical coverage indicates that the simulator captures stochastic variability of timeline lengths (see Methods). The coverage matching demonstrated only minimum errors (MAE range 0.0616–0.0180) across all time horizons. MAE, mean absolute error; KS, Kolmogorov–Smirnov. MAEs and KS statistics are shown with 95% confidence intervals.

The simulator generated high-fidelity clinical laboratory test data (Fig. 4 and Extended Data Fig. 3). We first assessed co-occurrence of the major laboratory tests. The real and simulation co-occurrence matrices showed visually identical patterns (Fig. 4a), with good agreement across all time horizons (r > 0.999; Fig. 4b and Extended Data Fig. 3a). We then evaluated the correlations between laboratory test result values. The real and

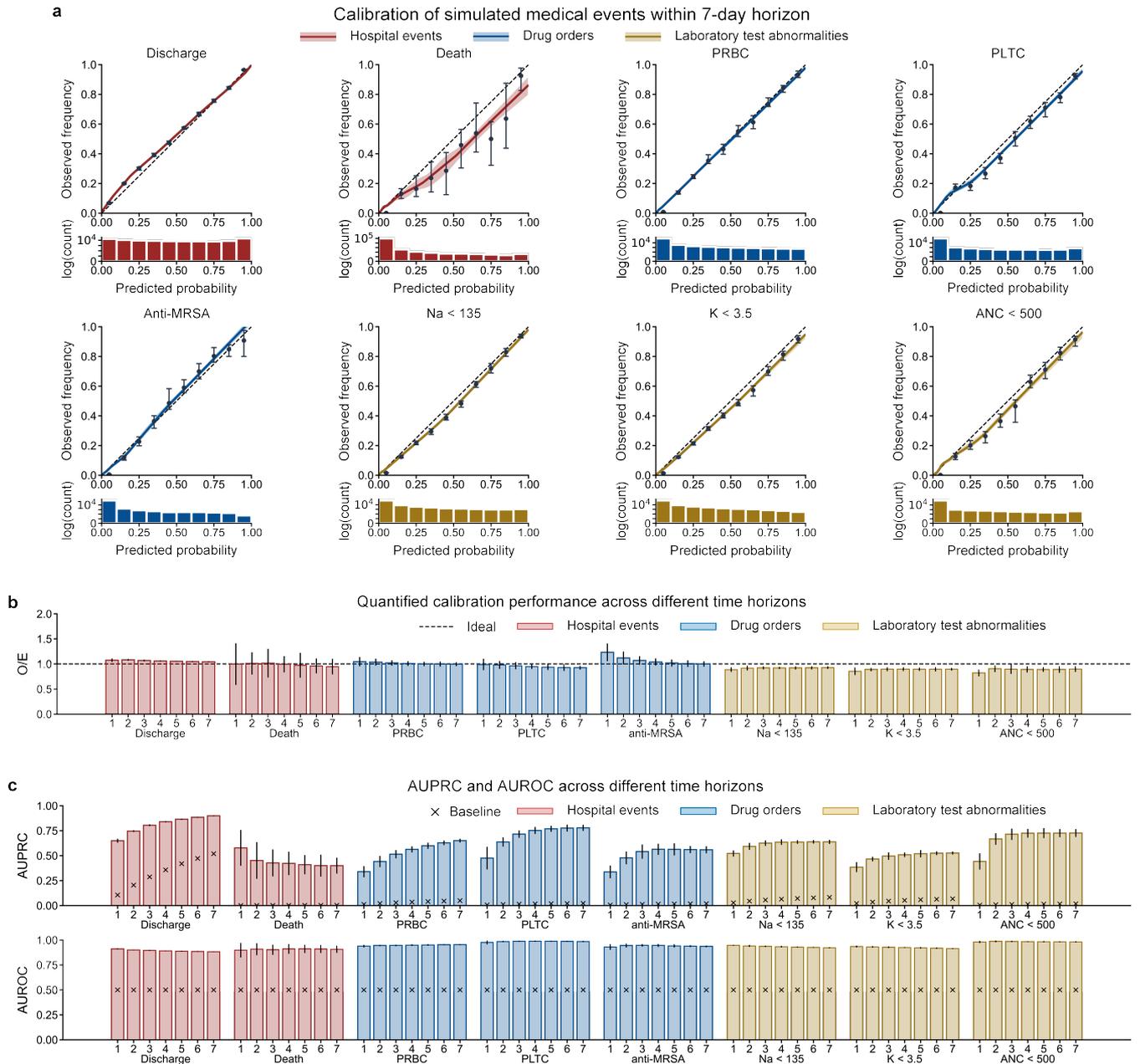

**Fig. 6 | The pretrained model simulated various medical clinical events with realistic occurrence rates.** This figure summarizes the predictive performance of the Monte Carlo method for the selected prediction tasks. For all the selected medical events, the model exhibited good calibration as well as high discrimination and precision-recall performance. **a**, Calibration curves. Each curve is supplemented by 10 equally spaced calibration bins. The bottom histograms indicate the distribution of predicted probabilities. The 95% confidence interval is shown as a shaded area. All curves closely align with the diagonal, indicating accurate calibration. **b**, Quantitative calibration performance measured by the O/E ratio. All O/E ratios were close to 1.0 without post hoc calibration. **c**, AUPRC and AUROC for each event prediction. Both metrics exceeded random baselines, indicating that the model assigns appropriate likelihoods to future events given the patient's context. ANC, absolute neutrophil count; AUPRC, area under the PR curve; AUROC, area under the ROC curve; CI, confidence interval; MRSA, methicillin-resistant *Staphylococcus Aureus*; O/E ratio, observed-to-expected ratio; PLTC, platelet concentrates; PR, precision-recall; PRBC, packed red blood cells; ROC, receiver operating characteristic.

simulation correlation matrices for major numeric laboratory tests also exhibited highly concordant structures (Fig. 4c), with strong agreement across time horizons (r > 0.980; Fig. 4d; Extended Data Fig. 3b). The simulator also accurately reproduced the marginal distributions of individual numeric laboratory test results. Density histograms of the simulated values closely matched those of the real data, capturing a wide range of distribution patterns, including normal-like, log-normal, and discrete-like distribution shapes (Fig. 4e). We measured distributional similarity using the Kolmogorov–Smirnov (KS) statistic. The KS statistic ranges from 0 to 1, with

0 indicating a perfect match. The generated laboratory test data demonstrated low KS statistics. At the 7-day horizon, the median KS statistic was 0.01785 (95% CI: 0.01529–0.02626; Fig. 4f), with similar close fits observed across other time horizons (Extended Data Fig. 3c). The simulator consistently outperformed the baseline model across all fidelity metrics (co-occurrence, correlation, and distributional similarity) at every time horizon. A list of the included numeric laboratory tests is provided in Supplementary Table 2. The list sufficiently covers routinely performed in vitro diagnostics.

The simulator accurately captured the temporal dynamics of patient timelines (Fig. 5). The distribution of timeline lengths in the simulated data closely matched those of the real data, with low KS statistics at the 7-day horizon (95% CI0.02350 ± 0.00140; Fig. 5a). The distribution of the number of temporal entries per timeline was also well reproduced (Fig. 5b). In addition to distributional similarity, the simulator exhibited well-calibrated stochastic variability in timeline length. The nominal coverage estimates derived from the simulated timelines closely matched the empirical coverage observed in the real data across all the time horizons. At the 7-day horizon, the mean absolute error between nominal and empirical coverage was 0.01798 (95% CI: 0.01678–0.01915).

In summary, the simulation framework demonstrated high-fidelity generative behaviour across medical event rates, laboratory test data, and temporal dynamics. In the following section, we evaluate the model's ability to predict medical events.

## Event risk prediction via the Monte Carlo method

We evaluated the model using Monte Carlo simulation to assess its ability to predict clinically meaningful outcomes. Although the model can, in principle, generate any event represented in its vocabulary, exhaustive evaluation is infeasible. We therefore focused on eight clinically important tasks spanning different event types and prediction challenges. These included two hospital outcomes (discharge and in-hospital death). In addition, three treatment events, packed red blood cell transfusion, platelet concentrate transfusion, and anti-MRSA (methicillin-resistant Staphylococcus aureus) drug orders, and three laboratory abnormalities (hyponatraemia, hypokalaemia, and neutropaenia) were included. These events are associated with serious complications and require nuanced modelling of individual patient trajectories. See Methods for definitions and rationale[20–27].

The simulator was well calibrated across all tasks and time horizons (Fig. 6). Events other than hospital discharge were infrequent, as illustrated by the event histograms beneath each calibration curve (Fig. 6a). The calibration curves for the 7-day prediction horizon showed close alignment with the ideal diagonal across all events. This calibration pattern was consistent across other time horizons (Supplementary Fig. 5). The observed-to-expected (O/E) ratios were close to 1.0 for all the tasks at every horizon (Fig. 6b), indicating that the probabilities predicted from the Monte Carlo simulations closely matched the real-world event frequencies.

Model performance was further supported by discrimination and precision-recall metrics. The area under the receiver operating characteristic curve (AUROC) and the area under the precision-recall curve (AUPRC) are summarized in Fig. 6c. Across all tasks and time horizons, the simulator consistently outperformed random baselines, often by large margins.

## Discussion

We developed a framework for simulating longitudinal patient trajectories and validated it in a real-world setting. The pretrained model learned from more than 200 million RWD records to generate realistic patient timelines. The simulated timelines demonstrated high fidelity across event rates, laboratory test data, and temporal dynamics. In these fidelity metrics, the model consistently outperformed the noncontextual empirical baseline. In addition, the model accurately predicted clinical events, producing probability estimates that closely matched observed event frequencies across diverse prediction tasks and time horizons. Our results show that autoregressive decoding applied to RWD enables empirical modelling of complex longitudinal clinical courses. This framework introduces a new foundation for longitudinal in silico patient modelling.

The strength of our framework is its flexibility in temporal generation. Existing clinical data generators have mainly used variational autoencoders, generative adversarial networks (GANs), and diffusion models[19,28]. These algorithms were originally developed for grid-like data such as images[29–31]. Therefore, generating clinical sequences of variable length and irregular timing is challenging. In contrast, patient timelines are inherently variable in both duration and temporal intervals. To address these limitations, we adopted an autoregressive framework based on transformer decoders[16]. Our model generates sequences of variable lengths and decodes time with minute-level resolution. We demonstrated that this approach can accurately be employed to simulate the temporal dynamics of future timelines (Fig. 5). Notably, we evaluated the model's performance on future patient trajectories, marking a substantial advance over prior studies that focused on reproducing historical data[19,28,32,33].

The framework also supports flexible simulation across both time horizons and clinical outcomes. Traditional supervised models are constrained to fixed prediction windows and a predefined set of outcomes (e.g., death in 30 days or 1 year)[34,35]. In contrast, our model allows arbitrary simulation horizons, including generation until a specified clinical event, such as hospital discharge, and estimates the probability of any future events represented in its vocabulary. This flexibility enables broad and dynamic exploration of patient trajectories. We also demonstrated that the Monte Carlo method based on our framework yielded well-calibrated, realistic probability estimates (Fig. 6). Together, the results of this study reveal that the autoregressive decoding approach, which is equipped with dedicated numeric and temporal encoders and decoders, enables high-fidelity, flexible simulations of longitudinal clinical courses.

Another strength of our framework is its ability to generate high-resolution laboratory test data. It has been challenging to synthesize numeric laboratory test values owing to their quasicontinuous nature. For example, haemoglobin is typically reported to one decimal place, whereas pH is reported to three (e.g., 12.5 mg/dL versus 7.423). Another challenge is the variable scales: white blood cell counts average approximately 4,000, whereas serum creatinine is near 1.0. Therefore, prior EHR data synthesis or simulation studies have typically excluded laboratory results[17,28]. Because laboratory test results account for a large proportion of clinical records, omitting them limits the representation and utility of simulations. In contrast, our model generates numeric values with clinical-level precision and appropriate scale and units. Our approach demonstrated high fidelity in multiple aspects of laboratory tests (Fig. 4). Another strength of our model is the broad range of tests that it covers. We demonstrated the model's high-fidelity behaviour in diverse laboratory tests, which accounted for 90% of the tests used in real-world practice. These tests sufficiently cover general routine laboratory tests[36]. Our method significantly advances the state of clinical data synthesis and simulation.

This framework is further enhanced by explainability, which is essential in high-stakes domains such as clinical medicine. However, most deep learning models operate as black boxes, offering little insight into how predictions are made. This lack of transparency can hinder clinical trust[37,38]. Our model provides two layers of explainability. First, it generates full patient timelines, allowing users to examine simulated trajectories in detail. For example, clinicians can review the simulated course of a patient who dies and assess likely causes or clinical pathways. Second, the model supports visualization of its inference process for each timeline entry, as shown in Fig. 2c. The visualized importance weights are the attention weights[39]. These weights are intermediate outputs produced during inference. Therefore, the weights provide explainability without additional computations. These features make our approach transparent.

Another important aspect of our study is that it highlights a new potential for RWD. While RWD offer invaluable resources, evidence-based medicine (EBM) has traditionally relied on prospectively collected data such as those from clinical trials. As a result, RWD have played a limited role in shaping EBM[3]. However, one critical advantage of RWD is its scale, which aligns well with the data demands of generative AI and foundation models. The major drawback of these models is that they require massive datasets for training[40]. Such volumes are infeasible to obtain through prospective studies alone. In this context, RWD are uniquely positioned to support the development of foundation models. In our study, we demonstrated that more than 460 million clinical records can be obtained from even a single hospital. Rather than selecting specific cohorts, we used the entire dataset to train a general-purpose clinical simulator. This approach demonstrates how RWD can be used at full scale. Further expansion is possible through widely adopted data standards such as HL7 and i2b2[11,12]. Our approach aligns with broader initiatives for large-scale RWD reuse, including the European Health Data Space and National Health Service Digital[14,41]. This work presents a new model for clinical research: using RWD as a scalable foundation for model-driven clinical science.

Finally, our framework can offer valuable opportunities. First, it can simply serve as a generator of synthetic medical data. Synthetic data have fuelled billion-dollar markets in other fields, and similar impacts are anticipated in clinical medicine[42]. Our method contributes to this space by producing granular, realistic, and privacy-preserving data. Our framework accepts any customized prompts. Therefore, users can generate patient data that fit their own needs by controlling prompts. The simulator can also support personalized medicine. When integrated with EHR, the simulator can act as a multipurpose patient forecasting engine through Monte Carlo estimation. A prototype implementation is shown in the Supplementary Video (the Supplementary Video is available at: https://doi.org/10.5281/zenodo.18357899). Another opportunity lies in hypothetical simulation. Because the model accepts structured tabular inputs, researchers can modify patient attributes or treatments to explore diverse clinical scenarios. This capability enables exploration across patient profiles for hypothesis generation and across treatment strategies to assess outcomes. These capabilities may support in silico clinical trials. Drug development can cost between 314 million and 2.8 billion US dollars, with clinical trial failure rates exceeding 50 percent[43,44]. Our model may offer a low-cost, early-stage tool to assess trial effectiveness before real-world studies, although its application for this purpose may require our proposed method in conjunction with rigorous causal inference techniques. Our framework can provide new opportunities for simulation-based medicine.

We acknowledge several limitations. First, the framework was evaluated at a single institution. Broader validation across diverse clinical settings is essential for establishing generalizability. Nonetheless, the model was tested under rigorous conditions: 103,579 patient trajectories were simulated 26,516,222 times under a strict temporal split, introducing temporal domain shifts and making validation more challenging[45]. Such testing is not feasible with publicly available datasets such as MIMIC-IV, where timestamps are perturbed for de-identification[46]. Second, the field of AI-based longitudinal patient simulation is still emerging. Existing models are scarce and differ substantially in supported modalities[17,18], making direct comparisons infeasible. Future studies should focus on benchmarking emerging methods on shared tasks using standardized formats. Third, we focused on short- to medium-term in-hospital trajectories. Outpatient settings require longer simulation horizons and pose additional challenges because of greater variability in care pathways. To ensure clinical consistency, we limited the evaluation to hospitalized patients and timeframes that span typical inpatient stays. Fourth, there are other types of medical records than those used in this study, such as vital signs. These records may contribute to more meaningful simulations. Although we did not include such records, our model architecture can natively support them. For example, vital signs can be integrated into our model in the same way as laboratory tests and their results are. Finally, we did not perform extensive hyperparameter optimization. Model performance, particularly in numeric value prediction, may improve with further tuning. For instance, we observed a slight overestimation of laboratory test abnormalities (Fig. 6), which may reflect suboptimal numeric decoding settings. Despite these limitations, we designed a comprehensive and challenging evaluation strategy to rigorously assess model performance.

In conclusion, we present a framework for longitudinal patient simulation that leverages RWD to generate fine-grained, high-fidelity patient timelines. The framework supports flexible simulations across variable lengths and time horizons and diverse clinical events and offers inherent explainability. We released the framework as open-source software to support continued research and broader adoption. This work lays the foundation for scalable, data-driven in silico patient modelling. It represents a step towards simulation-based medicine.

# References

1. Sun, P. et al. Scalability in perception for autonomous driving: Waymo open dataset in *Proceedings of the IEEE/CVF Conference on Computer Vision and Pattern Recognition* 2446–2454 (IEEE, 2020).
2. Jumper, J. et al. Highly accurate protein structure prediction with AlphaFold. *Nature* 596, 583–589 (2021).
3. Subbiah, V. The next generation of evidence-based medicine. *Nat. Med.* 29, 49–58 (2023).
4. Katsoulakis, E. et al. Digital twins for health: a scoping review. *NPJ Digit. Med.* 7, 77 (2024).
5. Creemers, J. H. A. et al. In silico cancer immunotherapy trials uncover the consequences of therapy-specific response patterns for clinical trial design and outcome. *Nat. Commun.* 14, 2348 (2023).
6. Prosperi, M. et al. Causal inference and counterfactual prediction in machine learning for actionable healthcare. *Nat. Mach. Intell.* 2, 369–375 (2020).
7. Desai, R. J. et al. Individualized treatment effect prediction with machine learning—salient considerations. *NEJM Evid.* 3, EVIDoa2300041 (2024).
8. Yang, L., Yang, J., Kleppe, A., Danielsen, H. E. & Kerr, D. J. Personalizing adjuvant therapy for patients with colorectal cancer. *Nat. Rev. Clin. Oncol.* 21, 67–79 (2024).
9. Rajkomar, A., Dean, J. & Kohane, I. Machine learning in medicine. *N. Engl. J. Med.* 380, 1347–1358 (2019).
10. Yoon, J. et al. EHR-safe: generating high-fidelity and privacy-preserving synthetic electronic health records. *NPJ Digit. Med.* 6, 141 (2023).
11. Kohane, I. S., Churchill, S. E. & Murphy, S. N. A translational engine at the national scale: informatics for integrating biology and the bedside. *J. Am. Med. Inform. Assoc.* 19, 181–185 (2012).

12. Wagholikar, K. B. et al. SMART-on-FHIR implemented over i2b2. *J. Am. Med. Inform. Assoc.* 24, 398–402 (2017).
13. Nakayama, M., Hui, F. & Inoue, R. Coverage of clinical research data retrieved from standardized structured medical information eXchange storage. *Stud. Health Technol. Inform.* 290, 3–6 (2022).
14. Raab, R. et al. Federated electronic health records for the European health data space. *Lancet Digit. Health* 5, e840–e847 (2023).
15. Sun, Z. et al. Fast structured decoding for sequence models. *Adv. Neural Inf. Process. Syst.* 32 (2019).
16. Brown, T. et al. Language models are few-shot learners. *Adv. Neural Inf. Process. Syst.* 33, 1877–1901 (2020).
17. Kraljevic, Z. et al. Foresight—a generative pretrained transformer for modelling of patient timelines using electronic health records: a retrospective modelling study. *Lancet Digit. Health* 6, e281–e290 (2024).
18. Renc, P. et al. Zero shot health trajectory prediction using transformer. *NPJ Digit. Med.* 7, 256 (2024).
19. Yan, C. et al. A multifaceted benchmarking of synthetic electronic health record generation models. *Nat. Commun.* 13, 7609 (2022).
20. Lacroix, J. et al. Age of transfused blood in critically ill adults. *N. Engl. J. Med.* 372, 1410–1418 (2015).
21. Slichter, S. J. et al. Dose of prophylactic platelet transfusions and prevention of hemorrhage. *N. Engl. J. Med.* 362, 600–613 (2010).
22. Turner, N. A. et al. Methicillin-resistant *Staphylococcus aureus*: an overview of basic and clinical research. *Nat. Rev. Microbiol.* 17, 203–218 (2019).
23. Carson, J. L. et al. Red blood cell transfusion: 2023 AABB international guidelines. *JAMA* 330, 1892–1902 (2023).
24. Metcalf, R. A. et al. Platelet transfusion: 2025 AABB and ICTMG international clinical practice guidelines. *JAMA* 334, 606–617 (2025).
25. Chen, S., Jalandhara, N. & Batlle, D. Evaluation and management of hyponatremia: an emerging role for vasopressin receptor antagonists. *Nat. Clin. Pract. Nephrol.* 3, 82–95 (2007).
26. Unwin, R. J., Luft, F. C. & Shirley, D. G. Pathophysiology and management of hypokalemia: a clinical perspective. *Nat. Rev. Nephrol.* 7, 75–84 (2011).
27. Freifeld, A. G. et al. Clinical practice guideline for the use of antimicrobial agents in neutropenic patients with cancer: 2010 update by the infectious diseases society of America. *Clin. Infect. Dis.* 52, e56–e93 (2011).
28. Loni, M., Poursalim, F., Asadi, M. & Gharehbaghi, A. A review on generative AI models for synthetic medical text, time series, and longitudinal data. *NPJ Digit. Med.* 8, 281 (2025).
29. Goodfellow, I. J. et al. Generative adversarial nets. *Adv. Neural Inf. Process. Syst.* 27 (2014).
30. Kingma, D. P. & Welling, M. Auto-encoding variational Bayes in *2nd International Conference on Learning Representations* 14 (ICLR, 2014).
31. Ho, J., Jain, A. & Abbeel, P. Denoising diffusion probabilistic models. *Adv. Neural Inf. Process. Syst.* 33, 6840–6851 (2020).
32. Li, J., Cairns, B. J., Li, J. & Zhu, T. Generating synthetic mixed-type longitudinal electronic health records for artificial intelligent applications. *NPJ Digit. Med.* 6, 98 (2023).
33. Theodorou, B., Xiao, C. & Sun, J. Synthesize high-dimensional longitudinal electronic health records via hierarchical autoregressive language model. *Nat. Commun.* 14, 5305 (2023).
34. Subudhi, S. et al. Comparing machine learning algorithms for predicting ICU admission and mortality in COVID-19. *NPJ Digit. Med.* 4, 87 (2021).
35. Smith, L. A. et al. Machine learning and deep learning predictive models for long-term prognosis in patients with chronic obstructive pulmonary disease: a systematic review and meta-analysis. *Lancet Digit. Health* 5, e872–e881 (2023).
36. WHO Strategic Advisory Group of Experts on in Vitro Diagnostics. *The Selection and Use of Essential in Vitro Diagnostics: Report of the Fourth Meeting of the WHO Strategic Advisory Group of Experts on in Vitro Diagnostics, 2022 (including the Fourth WHO Model List of Essential in Vitro Diagnostics). WHO Technical Report Series No. 1053* (World Health Organization, 2023).
37. Reddy, S. Explainability and artificial intelligence in medicine. *Lancet Digit. Health* 4, e214–e215 (2022).
38. Ghassemi, M., Oakden-Rayner, L. & Beam, A. L. The false hope of current approaches to explainable artificial intelligence in health care. *Lancet Digit. Health* 3, e745–e750 (2021).

39. Vaswani, A. et al. Attention is all you need. *Adv. Neural Inf. Process. Syst.* 30 (2017).
40. Moor, M. et al. Foundation models for generalist medical artificial intelligence. *Nature* 616, 259–265 (2023).
41. Zhang, J. et al. Mapping and evaluating national data flows: transparency, privacy, and guiding infrastructural transformation. *Lancet Digit. Health* 5, e737–e748 (2023).
42. Abgrall, G., Monnet, X. & Arora, A. Synthetic data and health privacy. *JAMA* 333, 567–568 (2025).
43. Wouters, O. J., McKee, M. & Luyten, J. Estimated research and development investment needed to bring a new medicine to market, 2009-2018. *JAMA* 323, 844–853 (2020).
44. Hwang, T. J. et al. Failure of investigational drugs in late-stage clinical development and publication of trial results. *JAMA Intern. Med.* 176, 1826–1833 (2016).
45. Van Der Meijden, S. L. et al. Development and validation of artificial intelligence models for early detection of postoperative infections (PERISCOPE): a multicentre study using electronic health record data. *Lancet Reg. Health* 49, 101163 (2025).
46. Johnson, A. E. W. et al. MIMIC-IV, a freely accessible electronic health record dataset. *Sci. Data* 10, 1 (2023).

# Methods
## Datasets
We prepared a dataset to evaluate our simulation framework in a real-world setting. The dataset consists of clinical records from January 1st, 2011, to December 31st, 2023. We extracted clinical records from the University of Tokyo Hospital's standardized data storage. This storage used HL7 version 2.5 messages standardized by the Japanese national standard (standardized structured medical information eXchange storage, SS-MIX2)[13]. This standard enforces strict rules for storing clinical records. We extracted patient demographics, admission and discharge records, diagnoses, medication orders, and laboratory test results.

We mapped the diagnosis codes to the 10th Revision of the International Classification of Diseases (ICD-10) and the medication codes to the Anatomical Therapeutic Chemical Classification System (ATC), both of which are maintained by the World Health Organization (WHO)[47,48]. In contrast, laboratory tests rarely follow international standards. For example, well-known public EHR datasets such as MIMIC-III, MIMIC-IV, and the eICU Collaborative Research Database use locally defined laboratory test codes[46,49,50]. For this reason, we adopted the Japanese Laboratory Code Version 10 (JLAC10), a national standard[51], without mapping to other coding systems. Importantly, our framework is agnostic to coding systems because it processes medical codes through embeddings learned from the training data. This approach enables compatibility with any coding system, including both standardized and locally defined codes.

We excluded records missing critical components, including missing codes, timestamps required to determine temporal ordering, and laboratory test results without recorded values (Supplementary Fig. 1). Additional data preprocessing procedures are described in the Supplementary Information.

We included all patients who had at least one record of admission, discharge, medication order, or laboratory test results. We split the dataset into training, validation, and test sets at the patient level. First, we held out the test dataset through temporal splitting. Patients were ordered by their first recorded date in the clinical records. The newest 7% of patients who had at least one record in 2023 were reserved for the test set. The remaining patients were then randomly split into 98% for the training set and 2% for the validation set. Clinical records from after December 31st, 2022, were excluded from the development dataset so that the model could be trained using data up to 2022.

## Data preprocessing
We preprocessed clinical records to timelines. First, we defined key timestamps to determine the positions in the timeline. A record may be associated with multiple timestamps. For example, a diagnosis may include the time of onset and the time of diagnosis. Using the onset time could incorrectly place the diagnosis in a position before it was made. Therefore, we selected timestamps that could minimize such risk (i.e., data leakage). For admissions and discharges, we used the exact times of admission and discharge. For medication orders, we used the time when the order became valid. For laboratory tests, we used the specimen sampling time. Considering delays between sampling and reporting, we excluded laboratory test results that had not been reported at the time of inference from prompts to mimic real-world application during the model evaluation.

We converted the timestamps to the elapsed time from the patient's date of birth by subtracting the date of birth from the target timestamp. Since this measure approximates the patient's age, we refer to it as 'age' in this study. The patient's birth time was set to 00:00 AM. Therefore, the calculated patient age at the hour and minute levels corresponded to the clock time.

After processing the timestamps, we formed a structured timeline (Extended Data Fig. 1). The timeline begins with the patient's first recorded age, followed by a token for the patient's sex. Each subsequent entry starts with the patient's age, with all entries sharing the same age placed sequentially beneath that entry. The laboratory test codes and their corresponding result values alternate, ensuring that each result immediately follows its associated test code. Discharge disposition tokens follow directly after the discharge event token.

We tokenized some categorical timeline entries. These categorical values include patient sex, admission and discharge events, and discharge dispositions. Diagnoses, medications, and laboratory tests are also categorical values. We designed special tokens for predefined categorical entities (Supplementary Table 1). Since positive (+), negative (−), and equivocal (+/−) are among the most frequent nonnumeric laboratory test results, we predefined special tokens for them.

For all other nonnumeric laboratory test results, we generated dedicated tokens during training data preprocessing. Tokenization was performed based on the basis of their frequency of occurrence in the training data. For each laboratory test code, we ranked unique nonnumeric results in descending order of frequency. Afterwards, we assigned tokens accordingly. For example, the ABO blood type can be A, B, O, or AB. If A is the most frequently observed result, followed by O, then B, and finally AB, we assign the token [TOP1] to A, [TOP2] to O, [TOP3] to B, and [TOP4] to AB. This approach helps limit the total number of generated tokens for nonnumeric laboratory test results.

Diagnoses, medications, and laboratory codes were tokenized using a subtoken approach. For example, the ATC code for ceftriaxone, an antibiotic, is J01DD04. This code can be decomposed into a set of subtokens: [[J**], [01****], [D], [**D], [*04]]. Each subtoken is embedded in the model. We employed this approach to reduce the number of trainable embeddings, thereby improving training efficiency. Decomposition is controlled by hyperparameters. Our framework allows each unique code to be treated as a single subtoken (e.g., [J01DD04] for ceftriaxone). With this flexible medical code tokenization, our framework can be adapted to any coding system. In this study, we decomposed medical codes in accordance with the hierarchical structure of the original coding systems.

We prepared a subtoken representing each record type: [DX] for diagnosis codes, [MED] for medication codes, and [LAB] for laboratory test codes. If a diagnosis was provisional, the subtoken [PRV] was appended. For example, the provisional diagnosis of bacterial pneumonia (J159 in the ICD-10) can be taken as [[DX], [J****], [*15**], [***9*], [PRV]].

All timeline entries were vectorized before being passed to the model. Each categorical entry was converted into a vector containing embedding indices. Numeric laboratory test results were transformed into percentiles using all the laboratory test data in the training dataset, scaled by dividing by 100, and encoded as one-dimensional vectors. Patient age was represented as a five-dimensional vector capturing years, months, days, hours, and minutes. These components were divided into 120, 12, 29, 23, and 59 for scaling. This list of vectors is the input to the model. Numeric laboratory test result values were normalized using percentile values obtained from the training dataset (see the model vocabulary section for details).

## Model vocabulary

The vocabularies for categorical entities were determined based on the basis of all categorical entities present in the dataset, as well as predefined special tokens. This vocabulary group included tokens representing nonnumeric laboratory test results. All unique medical codes observed in the training dataset were treated as distinct vocabulary entries. Diagnosis codes with provisional flags were considered separate from those without such flags.

The numerical laboratory test results were discretized based on the basis of their percentile values. Simply discretizing using equally spaced bins fails to capture clinically meaningful resolutions because pathological laboratory test values densely exist near extremities. To achieve finer resolution near extreme values, we applied a nonuniform discretization strategy that combined logit transform. First, a finite numeric space was defined by the logit function and a small constant $\varepsilon$. The finite space $r$ was defined as follows:

$$r \in [\text{logit}(\varepsilon), \text{logit}(1-\varepsilon)], \qquad \varepsilon = 10^{-7}$$

. Here, the logit function is as follows:

$$\text{logit}(a) = \log\left(\frac{a}{1-a}\right), \quad a \in \mathbb{R}$$

. Afterwards the continuous space $r$ was discretized into $N_{\text{bins}}$ segments. Let $\tilde{\mathbf{Z}} \in \mathbb{R}^{N_{bins}}$ be a vector containing discretized bins:

$$\tilde{\mathbf{Z}} = \{z_i | i = 1, \cdots, N_{bins}\}$$

. Each element $z_i$ is as follows:

$$z_i = \text{logit}(\varepsilon) + \frac{(i-1)\bigl(\text{logit}(1-\varepsilon) - \text{logit}(\varepsilon)\bigr)}{N_{bins} - 1}, \quad i = 1, \cdots, N_{bins}$$

. The $\tilde{\mathbf{Z}}$ was then mapped to the percentile rank space. Let $\mathbf{P} \in \mathbb{R}^{N_{bins}}$ be a vector containing the discretized percentile rank bins:

$$\mathbf{P} = \{p_i \in (\mathbf{0}, \mathbf{100}) | i = 1, \cdots, N_{bins}\}$$

. Each element $p_i$ is as follows:

$$p_i = 100 \times \text{expit}(z_i) = \frac{100}{1 + e^{-z_i}}, \quad i = 1, \cdots, N_{bins}$$

. For each unique combination of laboratory test $l$ and unit $u$, an ordered list of percentile values $\mathbf{V}_{l,u}$ is as follows:

$$\mathbf{V}_{l,u} = \{v_i | i = 1, \cdots, N_{bins}\} \in \mathbb{R}^{N_{bins}}$$

. $v_i$ corresponds to the $p_i$-th percentile value of the given combination of $l$ and $u$ observed in the training data. We used the nearest-rank method to compute percentile values to ensure that the model's numeric output and inputs were always associated with a member of the original data. In this study, we used $N_{bins} = 601$.

The pretrained model has $\mathbf{P}$ as its numeric vocabulary. During decoding, the model outputs a percentile rank $p_j \in \mathbf{P}$, which is mapped back to the original scale as follows:

$$\hat{x} = v_j \in \mathbf{V}_{l,u}$$

, where $\hat{x}$ is the predicted numeric laboratory test result value. The model determines $l$ from the preceding laboratory test entry in the timeline and selects $u$ as the most frequently observed unit for that test in the training set.

To normalize the numeric laboratory test result values in the input timeline, the model uses the values in $\tilde{\mathbf{Z}}$. The laboratory test result value $x \in \mathbb{R}$ is normalized to $\tilde{x} \in [0, 1]$ as follows:

$$\tilde{x} = \frac{z_{i*} - min(\tilde{\mathbf{Z}})}{max(\tilde{\mathbf{Z}}) - min(\tilde{\mathbf{Z}})}, \quad \text{where } i *= \arg\min_i |v_i - x|$$

. We used $z_{i*}$ instead of $p_{i*}$ because $p_{i*}$ may require high numerical accuracy where $p_{i*} \approx 0$ and $p_{i*} \approx 1$. Therefore, this normalization can avoid floating-point instability related to $p_{i*}$.

Another group of the model's vocabulary is time progression. This vocabulary group allows the model to decode fine-grained time steps in the generated timelines. Time progression was discretized to integrate temporal information into the model's vocabulary. We applied two discretization schemes to accommodate both short-term and long-term temporal dynamics. The first scheme addressed time intervals that did not span 24 hours, dividing them into uniform bins based on the basis of a fixed time step (Supplementary Fig. 6). We discretized a progression of less than 24 hours into 10-minute intervals, resulting in 144 unique tokens. These tokens correspond to intervals such as 0 to less than 10 minutes, 10 to less than 20 minutes, and continuing up to 1,430 to less than 1,440 minutes. For example, if a timeline entry occurred at 08:05 AM and the next entry was placed at 11:32 AM on the same day, the time progression was 207 minutes. Afterwards, this time progression is assigned to the bin representing 200–210 minutes of progression. Each discretized interval serves as a distinct entry in the model's vocabulary. The step size (10 minutes) is a hyperparameter of the model.

Time progressions beyond 24 hours were discretized using a separate scheme (Supplementary Fig. 6). First, the maximum progression was set. In this study, we established an upper limit of 1,800 days. Afterwards, the days were grouped into irregularly spaced bins. Progressions from 1 to 30 days were binned in 1-day intervals; progressions from 31 to 180 days were binned in 10-day intervals; progressions from 181 to 360 days were binned in 30-day intervals; and those from 361 to 1,800 days were binned in 180-day intervals. Each bin was further subdivided by clock hours using a fixed time step. In our study, this time step was set to 60 minutes, resulting in 24 hourly bins within each time progression category. For instance, if an event occurred 32 days after a reference point at a clock time of 3:15, it is assigned to the 31–40 day bin and further discretized into the 03:00 to less than 04:00 hourly bin. These discretization parameters are all hyperparameters.

## Model architecture

The architecture is illustrated in Fig. 1b. The model employs three encoders at its entry level. The categorical encoder consists of a single embedding layer to encode categorical variables. Embeddings for subtokens representing a categorical value are summed elementwise. The next encoder is the numeric encoder, which has two feed-forward layers and encodes numeric laboratory test values. Its input is a one-dimensional vector containing normalized percentiles. The last encoder is the temporal encoder, which encodes patient age. Its input is a five-dimensional vector representing normalized patient age. Aside from the difference in input size, the structure of the temporal encoder is identical to that of the numeric encoder. All encoded outputs are L2 normalized and interleaved in the order of positions in the original timeline. The final output is a matrix of sequence length ($l_{seq}$) by model dimension ($d_{model}$). An additional vector, referred to as the admission encoding, is added to the encoded vectors to represent the admission status of each timeline entry. This vector is a learnable vector, whose dimension is equal to $d_{model}$. The admission encoding is added elementwise to the encoded vectors representing entries recorded during admission. Finally, the encoded vectors are normalized using layer normalization.

The encoded inputs are then passed to the transformer layer stack. We used transformer blocks that are modified from the original transformer architecture and the learned positional encodings[52]. In addition, we used the Gaussian error linear unit as the activation function[53]. Each transformer block has one multihead attention layer and a feed-forward layer, with layer normalizations shifted before each of them. An additional layer normalization was added after the last transformer block[16]. The final output from this transformer layer stack was then passed to the prediction head to produce logits over the vocabulary. We set the number of attention heads ($h$) and the dimension of the attention keys and values ($d_k$ and $d_v$) to 16 and 64, respectively. We set up 12 decoder-only transformer blocks. We set the model's maximum sequence length ($d_{seq}$) to 2048. The hidden dimension of the feed-forward layers ($d_{ff}$) was set to 3072, and $d_{model}$ was set to 1024. We determined the set of hyperparameters on the basis of the existing study from GPT-3[16]. We employed Flash Attention 2 for efficient attention computation[54]. For attention visualization, the arithmetic mean of attention weights was used.

## Decoding

The model generated a timeline by autoregressive decoding, where a predicted output was appended to the input. This updated input was used as the next input to produce the next entry. At each inference, logits were transformed into confidence scores via the softmax function. The model sampled one decoding token from its vocabulary by sampling from the scores as probabilities. Numeric laboratory test results were first decoded as percentiles and then converted back to the original scale using the laboratory test code placed one position ahead. If a laboratory test was associated with multiple units, the most frequently observed unit in the training data was used during decoding. To decode time progression, one bin was selected from the discretized time progression space, and time progression was then sampled uniformly from the bin.

We applied weak rules for decoding to enforce the timeline structure by controlling logits. Specifically, the logit for discharge was set to negative infinity unless the patient had been admitted. Similarly, the logit for admission was set to negative infinity if the patient was already admitted. Logits for laboratory test results were ignored unless a laboratory test code preceded them. Consecutive time progression tokens were also prevented.

When the input reached $d_{seq}$, we applied a sliding-window approach with a stride of 128 to create space for new tokens. Autoregressive decoding then continued within this updated context. We employed key-value caching throughout decoding to accelerate generation[55].

## Model training

Model weights were initialized with a normal distribution:

$$\mathcal{N}(0, 0.02)$$

. Biases were initialized to zero, and layer normalization weights were set to one. The residual layer weights were further scaled by $1/n_{res}$, where $n_{res}$ denotes the number of residual layers[16].

We conducted the initial pretraining using all the training data. Timelines shorter than $d_{seq}$ were padded with empty rows. Timelines longer than $d_{seq}$ were resampled multiple times in proportion to their length. The number of resamplings was computed as follows:

$$n_{\text{resampling}} = \min\left(10, \left\lceil \frac{l_{seq}}{d_{seq}} \right\rceil\right)$$

. A patient was resampled up to 10 times at maximum, to prevent overfitting to patients with long trajectories. The resampled timeline was sliced to $d_{seq}$. A random sequence with the length of $d_{seq} - 2$ was first sliced and the first two rows of patient demographics were prepended to ensure that patient demographics were included in a slice.

The pretraining was scheduled for a maximum of 100 epochs. We used cross-entropy loss as the training criterion. The AdamW optimizer[56] was used, with $\beta_1 = 0.9$, $\beta_2 = 0.999$ and $\varepsilon = 10^{-8}$. We used a weight decay of 0.01. During the first 5% of the scheduled training, the learning rate (LR) linearly increased from 0 to 0.0003. LR was then gradually decreased using cosine annealing to $10^{-5}$ at 90% of the scheduled training and held constant thereafter. A dropout rate of 0.1 was applied to residual connections. The batch size started at 32 and was increased to 256 within the first 1% of training in steps of 32, 64, and 256. Early stopping was applied if the validation loss did not improve for 5 consecutive epochs.

After the initial pretraining stage, we conducted fine-tuning. This training was intended to fit the model to more recent medical practice. We trained the model so that only the timeline entries from the latest year (2022) could influence the update of the model weights. Fine-tuning was conducted with a constant learning rate of $10^{-5}$ and a constant batch size of 256. Other training settings followed those of the initial pretraining.

For both pretraining and fine-tuning, we used four NVIDIA A100 80 GB GPUs.

## Prompt timeline sampling

From the test dataset, we identified patients who experienced at least one complete hospitalization between January 1 and December 31, 2023. For each patient, we extracted paired samples comprising a prompt timeline and its corresponding future timeline. Sampling was performed once per day during hospitalization, up to day 30, at 12:00 PM. The prompt timeline (events up to the sampling point) was provided to the model as input, and the subsequent timeline (events after the sampling point) served as the ground truth for evaluation.

This sampling strategy captures a diverse range of clinical states, including acute illness at admission, recovery near discharge, and critical deterioration prior to death. Importantly, it mimics a plausible real-world deployment scenario: if the model were trained on data up to 2022 and deployed in 2023 to simulate hospitalized patients for possible futures at 12:00 PM each day, how well could it predict their future clinical trajectories?

We set a target time horizon and then ran 256 simulations per prompt. Afterwards, we use these generated trajectories for the downstream analyses. We used 7 different time horizons from 1-day to 7-day horizons.

## Fidelity analysis of the generated timelines

To evaluate the fidelity of the simulated timelines, we compared all the sampled future timelines with all the generated timelines. Because we ran 256 simulations per prompt, we obtained one set of real future timelines and 256 sets of simulated timelines. We compared the real set with each simulation set to yield 95% uncertainty intervals for the fidelity metrics. We conducted this step for all the time horizons. We evaluated the fidelity of medical event occurrence, the reproduction of laboratory test results, and temporal dynamics.

To evaluate fidelity, we established a baseline model for comparison. This baseline is a conceptual model that exactly reproduces the empirical distributions observed in the training data across all fidelity metrics.

Because it does not perform any contextual reasoning or conditional generation, this model represents a simple, nonparametric simulator that memorizes and resamples from the training distribution. Similar models have been widely adopted in prior work on synthetic EHR data generation as an upper-bound reference[19]. When the distributional shift is minimal, the model can approximate an oracle in terms of distributional similarity, despite its conceptual simplicity.

In our study, this baseline serves as a reference for the best possible performance that can be achieved by a model that ignores context. It is therefore used to benchmark the degree to which our model learns context-dependent generative patterns. Outperforming this baseline is critical: it demonstrates that our model does not simply memorize and replay historical patterns but instead generates patient timelines that are appropriately adapted to their clinical context. As such, fidelity beyond the baseline reflects the model's capacity to generalize to unseen data while respecting temporal and conditional structure, particularly under distributional shifts.

To compute baseline fidelity metrics, we sampled from the training data in a manner consistent with the evaluation setup. Because the model was fine-tuned using data from 2022 and to minimize the distributional shift between the baseline and the ground truth, we restricted baseline sampling to records from 2022. We first extracted timelines from this subset using the same prompt sampling procedure applied in the main evaluation.

For each fidelity metric, we constructed baseline samples by randomly selecting timelines from the 2022 subset without replacement. The number of sampled timelines matched the number of real timelines used for evaluation. Metrics such as medical code prevalence, timeline length distribution, and temporal entry counts were then computed on the sampled data. This process was repeated 256 times to obtain empirical distributions of the baseline performance, from which 95% confidence intervals were derived. This procedure ensured that baseline comparisons reflected the upper limit of noncontextual performance under equivalent sampling assumptions.

We evaluated the fidelity of medical event occurrence using the dimensionwise prevalence of medical codes. Because the model incorporates an extensive number of medical codes in its vocabulary, the evaluation focused on major medical codes. We defined the major codes as the most frequent medical codes that accounted for 90% of the total volume in the development dataset in each code category: diagnosis, medication and laboratory test. We directly compared the prevalence of these major codes in real versus simulation timeline sets.

Laboratory test results were analysed with additional metrics. We compared the co-occurrence of major laboratory tests within a time horizon in real versus simulated data. We also evaluated the correlations of numeric laboratory test results. Test-pair correlation was computed as Pearson's correlation coefficient using measurements that shared the same specimen sampling time. We excluded exceedingly rare test pairs (less than 0.5% of the total laboratory test orders) because the observations were insufficient for correlation analysis. We also evaluated the fidelity of the laboratory test result distributions. We compared the distribution of test result values from the major laboratory tests in real and simulation timeline sets. Distribution distance was measured using Kolmogorov–Smirnov statistics. For correlation and distribution analyses, we included only the laboratory tests that were associated with numeric result values in the test dataset because correlation and distribution could not be determined.

We evaluated fidelity for the temporal dynamics of timelines. We compared the distributions of both the timeline length and the number of temporal entries in the timelines in the real and simulation timeline sets. In addition to directly comparing these distributions, we also evaluated the model's ability to stochastically reproduce the true variability in timeline lengths. Specifically, we assessed the calibration of timeline length generation using nominal and empirical coverage. The nominal coverage of the timeline length refers to the percentile interval in the simulated samples that is expected to contain the real timeline length. For example, a 90% nominal coverage interval corresponds to the 5th–95th percentile of simulated lengths. The empirical coverage is the actual proportion of prompts for which the real timeline length falls within the corresponding nominal interval. We computed the empirical coverage across 21 nominal levels (0%, 5%, …, 95%, 100%). The discrepancy between the two was quantified using the mean absolute error. This evaluation assesses how accurately the model captures the stochastic distribution of timeline lengths beyond pointwise estimates. For this coverage evaluation, we used nonparametric bootstrapping to compute the 95% CI.

**Event prediction using the Monte Carlo method**

We evaluated how well the model could simulate clinically relevant medical events. We evaluated this capability as a prediction task. To estimate the event probability, we used the Monte Carlo method (Fig. 1c). The predicted event rate, $\hat{p}$, was computed as follows:

$$\hat{p} = \frac{n_{event}}{n_{sim}}$$

, where $n_{sim}$ denotes the total number of simulated timelines and $n_{event}$ is the number of those with positive events. In our study, $n_{sim} = 256$.

We evaluated the model performance on eight clinically important events spanning three task categories: hospital outcomes, medication orders, and laboratory abnormalities.

For hospital outcomes, we selected two universal and clinically critical endpoints: hospital discharge and in-hospital death. These outcomes are applicable to all hospitalized patients and are frequently used as primary endpoints in clinical trials.

For medication orders, we included orders for packed red blood cells, platelet concentrates, and anti-MRSA agents. These interventions are typically initiated in response to severe clinical complications such as anaemia, thrombocytopaenia, and MRSA infection. These complications may occur across a broad spectrum of hospitalized patients, and accurate forecasting of treatment decisions requires a nuanced understanding of patient backgrounds. At present, these clinical decisions are not routinely supported by machine learning-based tools, making them both clinically relevant and technically challenging for predictive models[20-24].

Finally, we assessed the prediction of laboratory test abnormalities, including hyponatraemia, hypokalaemia, and neutropaenia. These conditions are common in inpatient populations and are associated with increased morbidity and mortality[24,57-61]. Among them, neutropaenia is a particularly high-risk event that often follows chemotherapy and is difficult to anticipate because of its idiosyncratic nature. Risk assessment for these laboratory abnormalities currently relies heavily on clinical judgement, and robust predictive tools are lacking[25-27]. These tasks were therefore chosen to test the model's ability to simulate complex, high-variance physiological states.

To detect medication orders, we used ATC codes. Packed red blood cells and platelet concentrates were identified using B05AX01 and B05AX02, respectively. Anti-MRSA drug orders were defined using J01XX08 (linezolid), J01XX09 (daptomycin), and J01XA along with all of its child concepts (glycopeptides, including vancomycin). Laboratory test abnormalities were defined on the basis of established diagnostic thresholds. Hyponatraemia was defined as a serum sodium concentration < 135 mmol/L, hypokalaemia was defined as a serum potassium concentration < 3.5 mmol/L, and neutropaenia was defined as an absolute neutrophil count < 500/μL. These criteria are widely used for clinical diagnosis of the respective conditions[25-27].

We evaluated the calibration, discrimination and precision-recall performance of the model. For calibration, we plotted calibration curves using locally estimated scatterplot smoothing (LOESS), with 20% of the data points used for each local fit[62]. We plotted an empirical calibration using 10 equally spaced probability bins for reference. We quantified the calibration using the O/E ratio, which was computed as the total number of observed events divided by the total number of expected events[62]. For discrimination and precision-recall performance, we measured the AUROC and AUPRC and compared them with those of the random baseline. We used nonparametric bootstrapping to estimate the 95% CI.

## Ethics oversight

This study was approved by the institutional review boards of the University of Tokyo (2023367NI). The approval included patients of all ages, with informed consent obtained using an opt-out method. Patients or their representatives were provided a sufficient opportunity to decline participation through public notices. Individual written consent was waived because of the retrospective nature of the study.

## Data availability

The dataset and data generated by the trained model may contain pseudonymized information that could potentially reveal sensitive details about participants. To comply with Japan's Act on the Protection of Personal Information, including restrictions on the third-party provision of personal data, the dataset and model weights are not publicly available. For inquiries regarding reproducibility of the research, please contact the corresponding author.

## Code availability

All the codes are publicly available. The simulation model, along with the data generation and preprocessing scripts, is provided at https://yuakagi.github.io/Watcher/. The prototype digital-twin EHR web application is available at https://yuakagi.github.io/TwinEHR/. The data pipeline Python package for Japanese EHR storage used in our study is available at https://yuakagi.github.io/ssmixtools/.

## Methods References

47. World Health Organization. *International Classification of Diseases, 10th Revision (ICD-10)* (World Health Organization, 2019).
48. WHO Collaborating Centre for Drug Statistics Methodology. *Guidelines for ATC Classification and DDD Assignment 2025* (WHO Collaborating Centre for Drug Statistics Methodology, 2024).
49. Johnson, A. E. W. et al. MIMIC-III, a freely accessible critical care database. *Sci. Data* 3, 160035 (2016).
50. Pollard, T. J. et al. The eICU collaborative research database, a freely available multi-center database for critical care research. *Sci. Data* 5, 180178 (2018).
51. Kume, N., Suzuki, K., Kobayashi, S., Araki, K. & Yoshihara, H. Development of unified lab test result master for multiple facilities. *Stud. Health Technol. Inform.* 216, 1050 (2015).
52. Radford, A., Narasimhan, K., Salimans, T. & Sutskever, I. Improving language understanding by generative pre-training. *OpenAI* (2019).
53. Hendrycks, D. & Gimpel, K. Gaussian error linear units (GELUS). arXiv preprint arXiv:1606.08415 (2016).
54. Dao, T. Flashattention-2: faster attention with better parallelism and work partitioning. arXiv preprint arXiv:2307.08691 (2023).
55. Pope, R. et al. Efficiently scaling transformer inference in *Proceedings of Machine Learning and Systems* 606–624 (MLSys, 2023).
56. Loshchilov, I. & Hutter, F. Decoupled weight decay regularization in *7th International Conference on Learning Representations* (ICLR, 2019).
57. Yataco, A. O. C. et al. Red blood cell transfusion in critically ill adults: an American college of chest physicians clinical practice guideline. *Chest* 167, 477–489 (2025).
58. Popovich, K. J. et al. SHEA/IDSA/APIC practice recommendation: strategies to prevent methicillin-resistant *Staphylococcus aureus* transmission and infection in acute-care hospitals: 2022 update. *Infect. Control Hosp. Epidemiol.* 44, 1039–1067 (2023).
59. Spasovski, G. et al. Clinical practice guideline on diagnosis and treatment of hyponatraemia. *Nephrol. Dial. Transplant.* 29, i1–i39 (2014).
60. Kardalas, E. et al. Hypokalemia: a clinical update. *Endocr. Connect.* 7, R135–R146 (2018).
61. Fioredda, F. et al. The European guidelines on diagnosis and management of neutropenia in adults and children: a consensus between the European hematology association and the EuNet-INNOCHRON COST action. *HemaSphere* 7, e872 (2023).
62. Riley, R. D. et al. Evaluation of clinical prediction models (part 2): how to undertake an external validation study. *BMJ* 384, e074820 (2024).
63. Braat, S. et al. Haemoglobin thresholds to define anaemia from age 6 months to 65 years: estimates from international data sources. *Lancet Haematol.* 11, e253–e264 (2024).

## Acknowledgements

This work was supported by the Moonshot Research and Development Program (Cabinet Office, Government of Japan, grant number JP22zf0127006); Council for Science, Technology and Innovation (CSTI); the Cross-ministerial Strategic Innovation Promotion Program (SIP); the 3rd period of SIP, 'Integrated Health Care System' (Cabinet Office, Government of Japan, grant number JPJ012425); and Support for Pioneering Research Initiated by the Next Generation (Japan Science and Technology Agency, grant number JPMJSP2108). Y.A. received research support from the Saito Scholarship of the Japanese Society for Medical and Biological Engineering.

## Author contributions

Y.A. conceptualized the study. Y.A. drafted the initial manuscript and developed the data pipeline. K.O. managed the EHR servers, including data cloning and de-identification. Y.A., T.S., T.T., and H.I. mapped and cleaned clinical records. Y.A. coded the simulation model, trained and fine-tuned it, and conducted the performance evaluation. Y.K., T.S., T.T., and H.I. advised on statistical analysis. Y.K. and K.O. supervised the project.

## Competing Interest

Authors YK and KO belong to the Artificial Intelligence and Digital Twin in Healthcare, Graduate School of Medicine, University of Tokyo, which is an endowment department supported by an unrestricted grant from EM Systems, EPNextS, MRP Co., Ltd., SHIP HEALTHCARE HOLDINGS, Inc., SoftBank Corp., NEC Corporation, and Nippon Sogo Systems, Inc.; these organizations had no control over the interpretation, writing, or publication of this work.

# Extended Data Table 1 | Patient and timeline characteristics

|  | Development dataset (*n*=348,115) | | Test dataset (*n*=25,735) |
| --- | --- | --- | --- |
|  | Training cohort (*n*=340,659) | Validation cohort (*n*=7,496) | Test admission cohort (*n*=7,907) |
| **Target period** | Jan 2011–Dec 2022 | Jan 2011–Dec 2022 | Jan 2023–Dec 2023 |
| **Age** | | | |
| Median (years) | 50 (31–67) | 50 (32–67) | 54 (33–70) |
| 0–18 | 44,470 | 948 | 1,011 |
| 19–64 | 197,892 | 4,325 | 4,291 |
| ≥65 | 98,297 | 2,223 | 2,605 |
| **Sex** | | | |
| Female | 179,552 (52.7%) | 3,987 (53.2%) | 4,191 (53.0%) |
| Male | 161,137 (47.3%) | 3,509 (46.8%) | 3716 (47.0%) |
| **Whole timeline** | | | |
| Median duration (days) | 301 (28–1,618) | 304 (28–1,626) | 128 (54–243) |
| Median length (number of entries) | 195 (30–947) | 192 (29–925) | 653 (333–1578) |
| **Admission** | | | |
| Total admissions | 317,715 | 6,639 | 12,606 |
| In-hospital deaths | 3,748 (1.2%) | 84 (1.3%) | 102 (0.81%) |
| Median hospitalized days | 6 (3–14) | 7 (3–14) | 5 (2–12) |
| **Admission by clinical departments** | | | |
| Medical | 92,327 | 2,049 | 2,783 |
| Surgical | 82,552 | 1,689 | 3,604 |
| Pediatric | 17,495 | 258 | 1,180 |
| Obstetrics and gynecology | 41,177 | 822 | 1,624 |
| Critical and emergency care | 8,620 | 195 | 455 |
| Others | 75,544 | 1,626 | 2,960 |

Median values are shown with interquartile ranges in parentheses. The test admission cohort is a subset of the test dataset that includes only patients with at least one complete admission.

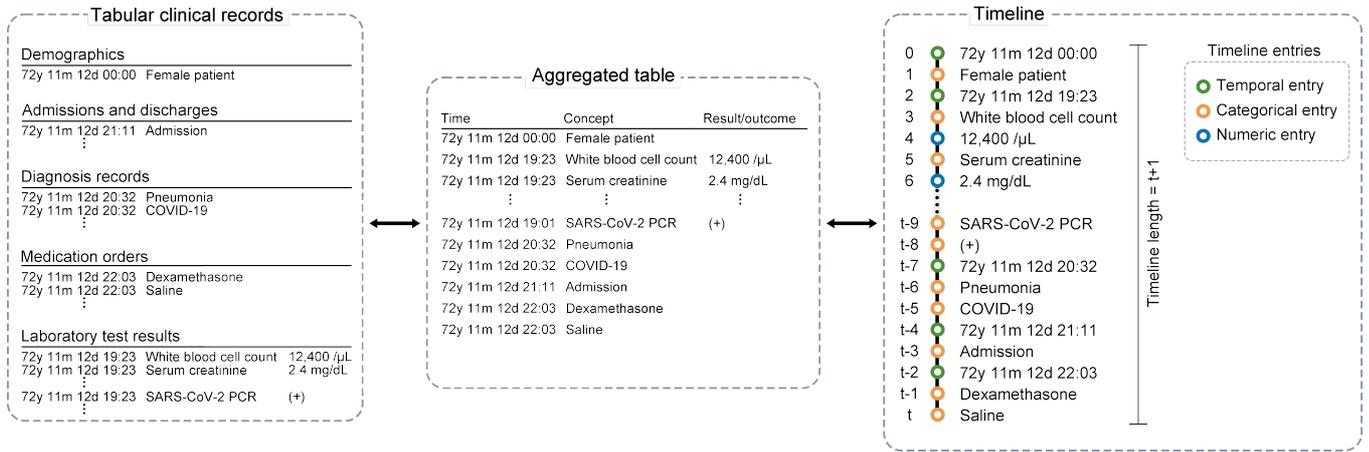

**Extended Data Fig. 1 | Structure of patient timelines.** Raw clinical records were initially formatted as multiple separate tables. These tables were aggregated into a unified table per patient and then transformed into a single timeline. Each element in the timeline is referred to as an entry. There are three types of entries: temporal (green), categorical (orange), and numeric (blue). A temporal entry defines a unique time point, and all clinical events occurring at that time are placed beneath it. Laboratory test results, either numeric or categorical, are positioned immediately after their corresponding test to preserve the association. Discharge disposition entries (e.g., survived or died) are placed immediately after the discharge event. The timeline length is defined as the total number of entries. The process is reversible, allowing timelines to be converted back into tabular format. Tabular formats were used for downstream analysis.

**a**

| Record type | Patient age | Code / token | Entry name | Value / status |
|---|---|---|---|---|
| laboratory test | 3 days 16:57:00 | 3H0800000019---51 | whole blood,blood gas analysis,pH, | 7.346 |
| laboratory test | 3 days 16:57:00 | 3H0800000019---52 | whole blood,blood gas analysis,pCO2, | 43.0 mmHg |
| laboratory test | 3 days 16:57:00 | 3H0800000019---53 | whole blood,blood gas analysis,pO2, | 36.4 mmHg |
| laboratory test | 3 days 16:57:00 | 3H0800000019---54 | whole blood,blood gas analysis,actual bicarbonate, | 22.8 mmol/L |
| laboratory test | 3 days 16:57:00 | 3H0800000019---55 | whole blood,blood gas analysis,base excess, | -2.5 mmol/L |
| laboratory test | 3 days 16:57:00 | 3H0800000019---56 | whole blood,blood gas analysis,oxygen saturation, | 76.6 % |
| laboratory test | 3 days 16:57:00 | 3H0800000019---58 | whole blood,blood gas analysis,hemoglobin, | 17.0 g/dL |
| laboratory test | 3 days 16:57:00 | 3H0800000019---59 | whole blood,blood gas analysis,hematocrit, | 52.0 % |
| laboratory test | 3 days 16:57:00 | 3H0800000019---81 | whole blood,blood gas analysis,ionized sodium, | 137.0 mmol/L |
| laboratory test | 3 days 16:57:00 | 3H0800000019---82 | whole blood,blood gas analysis,ionized pottasium, | 4.4 mmol/L |
| laboratory test | 3 days 16:57:00 | 3H0800000019---83 | whole blood,blood gas analysis,ionized chlorine, | 110.0 mmol/L |
| laboratory test | 3 days 16:57:00 | 3H0800000019---84 | whole blood,blood gas analysis,ionized calcium, | 1.27 mmol/L |
| laboratory test | 3 days 16:57:00 | 3H0800000019---85 | whole blood,blood gas analysis,glucose, | 85.0 mg/dL |
| laboratory test | 3 days 16:57:00 | 3H0800000019---86 | whole blood,blood gas analysis,lactate, | 2.0 mmol/L |
| drug order | 4 days 13:48:00 | B02BA | Vitamin K | |
| diagnosis | 4 days 16:21:00 | P616 | Other transient neonatal disorders of coagulation | |
| laboratory test | 4 days 17:30:00 | 3H0800000019---51 | whole blood,blood gas analysis,pH, | 7.346 |
| laboratory test | 4 days 17:30:00 | 3H0800000019---52 | whole blood,blood gas analysis,pCO2, | 44.0 mmHg |
| laboratory test | 4 days 17:30:00 | 3H0800000019---53 | whole blood,blood gas analysis,pO2, | 43.1 mmHg |
| laboratory test | 4 days 17:30:00 | 3H0800000019---54 | whole blood,blood gas analysis,actual bicarbonate, | 23.4 mmol/L |
| laboratory test | 4 days 17:30:00 | 3H0800000019---55 | whole blood,blood gas analysis,base excess, | -1.9 mmol/L |
| laboratory test | 4 days 17:30:00 | 3H0800000019---56 | whole blood,blood gas analysis,oxygen saturation, | 83.2 % |
| laboratory test | 4 days 17:30:00 | 3H0800000019---58 | whole blood,blood gas analysis,hemoglobin, | 15.0 g/dL |
| laboratory test | 4 days 17:30:00 | 3H0800000019---59 | whole blood,blood gas analysis,hematocrit, | 46.0 % |
| laboratory test | 4 days 17:30:00 | 3H0800000019---81 | whole blood,blood gas analysis,ionized sodium, | 139.0 mmol/L |
| laboratory test | 4 days 17:30:00 | 3H0800000019---82 | whole blood,blood gas analysis,ionized pottasium, | 4.4 mmol/L |
| laboratory test | 4 days 17:30:00 | 3H0800000019---83 | whole blood,blood gas analysis,ionized chlorine, | 107.0 mmol/L |
| laboratory test | 4 days 17:30:00 | 3H0800000019---84 | whole blood,blood gas analysis,ionized calcium, | 1.36 mmol/L |
| laboratory test | 4 days 17:30:00 | 3H0800000019---85 | whole blood,blood gas analysis,glucose, | 81.0 mg/dL |
| laboratory test | 4 days 17:30:00 | 3H0800000019---86 | whole blood,blood gas analysis,lactate, | 1.6 mmol/L |
| drug order | 4 days 20:50:00 | B02BA | Vitamin K | |
| discharge event | 5 days 10:42:00 | [DSC] | discharged | survived |

**b**

| Record type | Patient age | Code / token | Entry name | Value / status |
|---|---|---|---|---|
| laboratory test | 3 days 06:46:00 | 3H0800000019---51 | whole blood,blood gas analysis,pH, | 7.401 |
| laboratory test | 3 days 06:46:00 | 3H0800000019---52 | whole blood,blood gas analysis,pCO2, | 34.7 mmHg |
| laboratory test | 3 days 06:46:00 | 3H0800000019---53 | whole blood,blood gas analysis,pO2, | 53.1 mmHg |
| laboratory test | 3 days 06:46:00 | 3H0800000019---54 | whole blood,blood gas analysis,actual bicarbonate, | 21.1 mmol/L |
| laboratory test | 3 days 06:46:00 | 3H0800000019---55 | whole blood,blood gas analysis,base excess, | -2.5 mmol/L |
| laboratory test | 3 days 06:46:00 | 3H0800000019---56 | whole blood,blood gas analysis,oxygen saturation, | 89.9 % |
| laboratory test | 3 days 06:46:00 | 3H0800000019---58 | whole blood,blood gas analysis,hemoglobin, | 14.4 g/dL |
| laboratory test | 3 days 06:46:00 | 3H0800000019---59 | whole blood,blood gas analysis,hematocrit, | 44.1 % |
| laboratory test | 3 days 06:46:00 | 3H0800000019---81 | whole blood,blood gas analysis,ionized sodium, | 136.0 mmol/L |
| laboratory test | 3 days 06:46:00 | 3H0800000019---82 | whole blood,blood gas analysis,ionized pottasium, | 3.6 mmol/L |
| laboratory test | 3 days 06:46:00 | 3H0800000019---83 | whole blood,blood gas analysis,ionized chlorine, | 107.0 mmol/L |
| laboratory test | 3 days 06:46:00 | 3H0800000019---84 | whole blood,blood gas analysis,ionized calcium, | 1.27 mmol/L |
| laboratory test | 3 days 06:46:00 | 3H0800000019---85 | whole blood,blood gas analysis,glucose, | 73.0 mg/dL |
| laboratory test | 3 days 06:46:00 | 3H0800000019---86 | whole blood,blood gas analysis,lactate, | 1.5 mmol/L |
| drug order | 3 days 11:14:00 | B02BA | Vitamin K | |
| laboratory test | 4 days 06:31:00 | 3H0800000019---51 | whole blood,blood gas analysis,pH, | 7.374 |
| laboratory test | 4 days 06:31:00 | 3H0800000019---52 | whole blood,blood gas analysis,pCO2, | 39.0 mmHg |
| laboratory test | 4 days 06:31:00 | 3H0800000019---53 | whole blood,blood gas analysis,pO2, | 34.2 mmHg |
| laboratory test | 4 days 06:31:00 | 3H0800000019---54 | whole blood,blood gas analysis,actual bicarbonate, | 22.2 mmol/L |
| laboratory test | 4 days 06:31:00 | 3H0800000019---55 | whole blood,blood gas analysis,base excess, | -2.2 mmol/L |
| laboratory test | 4 days 06:31:00 | 3H0800000019---56 | whole blood,blood gas analysis,oxygen saturation, | 70.5 % |
| laboratory test | 4 days 06:31:00 | 3H0800000019---58 | whole blood,blood gas analysis,hemoglobin, | 14.6 g/dL |
| laboratory test | 4 days 06:31:00 | 3H0800000019---59 | whole blood,blood gas analysis,hematocrit, | 44.8 % |
| laboratory test | 4 days 06:31:00 | 3H0800000019---81 | whole blood,blood gas analysis,ionized sodium, | 138.0 mmol/L |
| laboratory test | 4 days 06:31:00 | 3H0800000019---82 | whole blood,blood gas analysis,ionized pottasium, | 5.0 mmol/L |
| laboratory test | 4 days 06:31:00 | 3H0800000019---83 | whole blood,blood gas analysis,ionized chlorine, | 110.0 mmol/L |
| laboratory test | 4 days 06:31:00 | 3H0800000019---84 | whole blood,blood gas analysis,ionized calcium, | 1.33 mmol/L |
| laboratory test | 4 days 06:31:00 | 3H0800000019---85 | whole blood,blood gas analysis,glucose, | 66.0 mg/dL |
| laboratory test | 4 days 06:31:00 | 3H0800000019---86 | whole blood,blood gas analysis,lactate, | 2.1 mmol/L |
| diagnosis | 5 days 14:13:00 | P616 | Other transient neonatal disorders of coagulation | |
| drug order | 5 days 14:13:00 | B02BA | Vitamin K | |
| drug order | 5 days 17:14:00 | B02BA | Vitamin K | |
| diagnosis | 6 days 00:00:00 | P284 | Other apnoea of newborn | |
| diagnosis | 6 days 00:00:00 | Q210 | Ventricular septal defect | |
| discharge event | 6 days 10:00:00 | [DSC] | discharged | survived |

**Extended Data Fig. 2 | The pretrained model generated fine-grained timeline data that closely resembled real-world clinical records.** This example shows a hospitalized newborn patient. The pretrained model generated a synthetic timeline using the patient's data up to 12:00 PM on the second hospital day as input and continued the simulation until discharge. **a**, AI-generated patient timeline near discharge. **b**, Corresponding real-world patient timeline. Both timelines are presented in the tabular format. Each row represents a structured clinical observation, with columns indicating the record type, patient age (displayed with day–hour–minute resolution), medical code or token, descriptive label, and associated outcome (e.g., a laboratory test result or discharge disposition). The AI-generated trajectory closely mirrors the real case. Laboratory test results are reported using the same units and the same level of decimal precision in both timelines. Blood gas analyses appear in both, with consistent units, precision, and value scales. Vitamin K (B02BA), which often administered to newborns to prevent haemorrhage, is ordered in both records. The same diagnosis code, P284 ("transient neonatal disorders of coagulation" in ICD-10), also appears in both records. Due to space constraints, only excerpts near discharge are shown. The full simulation output is provided in Supplementary Fig. 3. AI, artificial intelligence; ICD-10, the 10th Revision of the International Classification of Diseases.

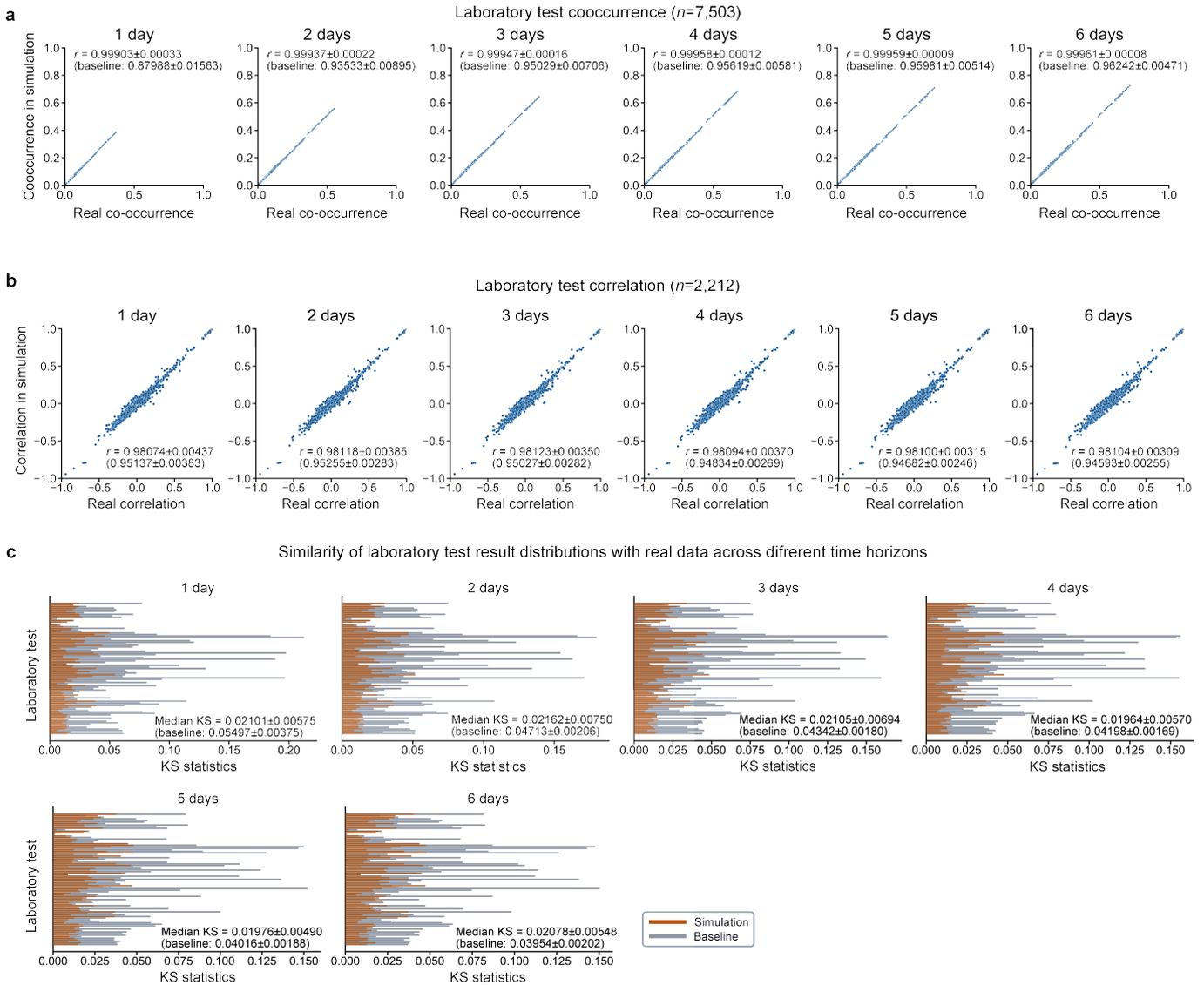

**Extended Data Fig. 3 | The simulator generated high-fidelity laboratory test data across different time horizons.** This figure shows the fidelity of simulated laboratory test data in different time horizons. Baseline performance is shown in parentheses. **a**, Agreement of major laboratory test co-occurrence between simulation and real data across different time horizons. Simulation data showed high agreement with the real data ($r > 0.999$). **b**, Agreement of major numeric laboratory test result correlations. Test-pair correlations were well reproduced in the simulation ($r > 0.980$). **c**, Similarity of numeric laboratory test result distributions between simulation and real. KS statistics were small in simulation across all time horizons. The mean KS statistic was lower than that of the baseline in all time horizons. KS, Kolmogorov–Smirnov.

# Supplementary Information

# High-Fidelity Longitudinal Patient Simulation Using Real-World Data

Yu Akagi[1*], Tomohisa Seki[2], Hiromasa Ito[2,3], Toru Takiguchi[2,4], Kazuhiko Ohe[5,6], Yoshimasa Kawazoe[2,5*]

[1] Department of Biomedical Informatics, Graduate School of Medicine, The University of Tokyo, Tokyo, Japan
[2] Department of Healthcare Information Management, The University of Tokyo Hospital, Tokyo, Japan
[3] Department of Cardiology and Nephrology, Mie University Graduate School of Medicine, Mie, Japan
[4] Department of Emergency and Critical Care Medicine, Nippon Medical School, Tokyo, Japan
[5] Artificial Intelligence and Digital Twin in Healthcare, Graduate School of Medicine, The University of Tokyo, Tokyo, Japan
[6] Graduate School of Health Data Science, Juntendo University, Tokyo, Japan

1 # Supplementary results

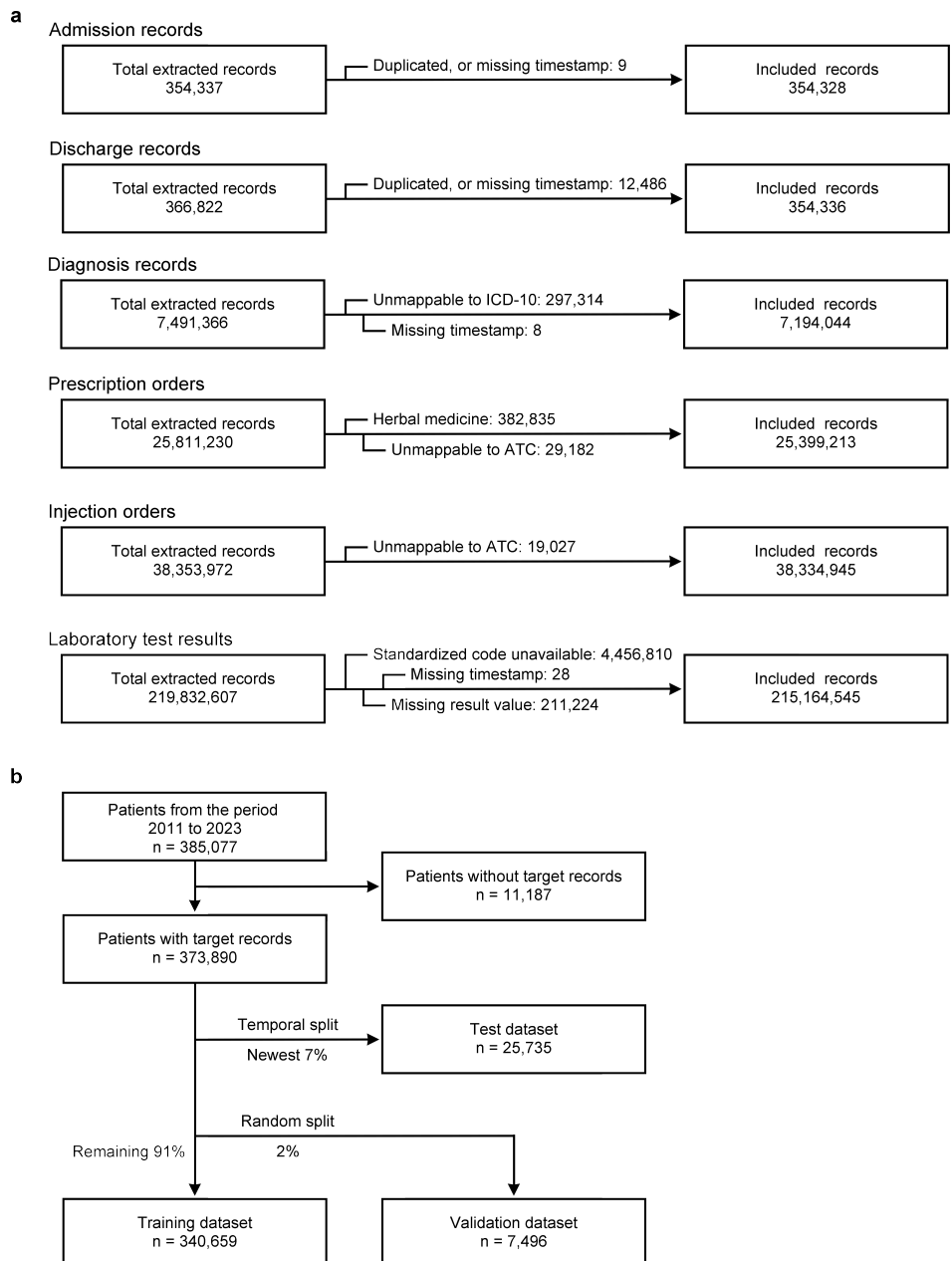

**Supplementary Fig. 1 | Overview of data selection and patient cohort selection. a**, Summary of record exclusion and retained record counts by record type. Records were excluded due to duplication, missing timestamps, or mapping failures. Details of the data cleaning procedure are provided in the Supplementary Information. **b**, Patient cohort selection flow. We included patients who had at least one record from any of the following categories: medication orders, laboratory test results, or admission history. Patients with none of these records were excluded. Patients were temporally aligned based on the date of their earliest available record. To construct the test dataset, we reserved the newest 7% of patients according to this alignment. The remaining patients were used for model development, including training and validation. ATC, Anatomical Therapeutic Chemical Classification System; ICD-10, the 10th Revision of the International Classification of Diseases.

2
3

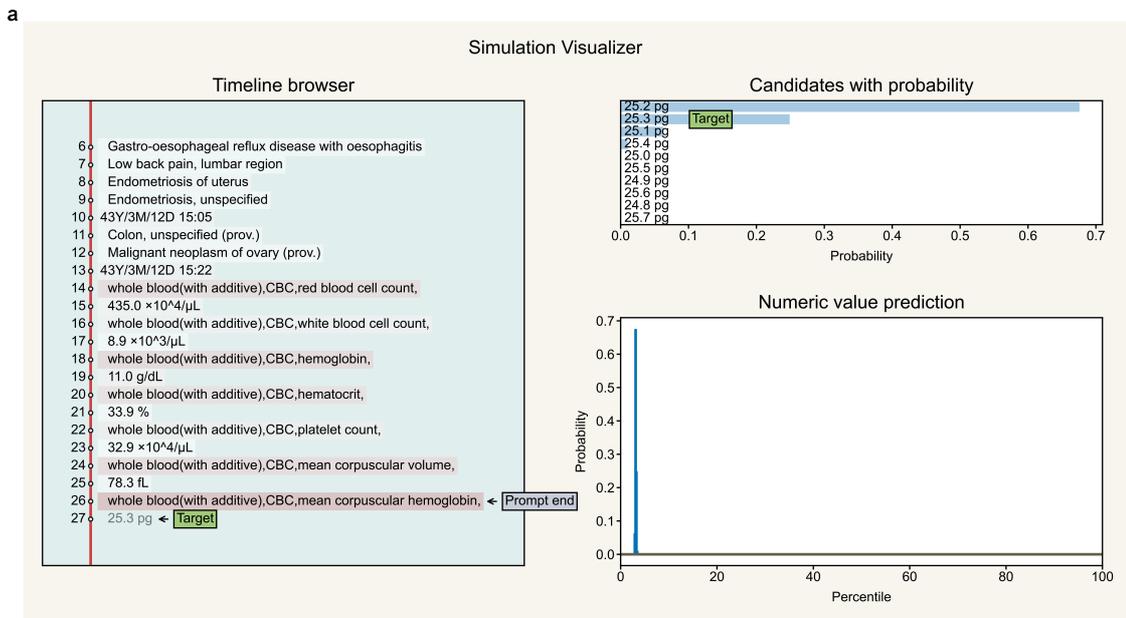

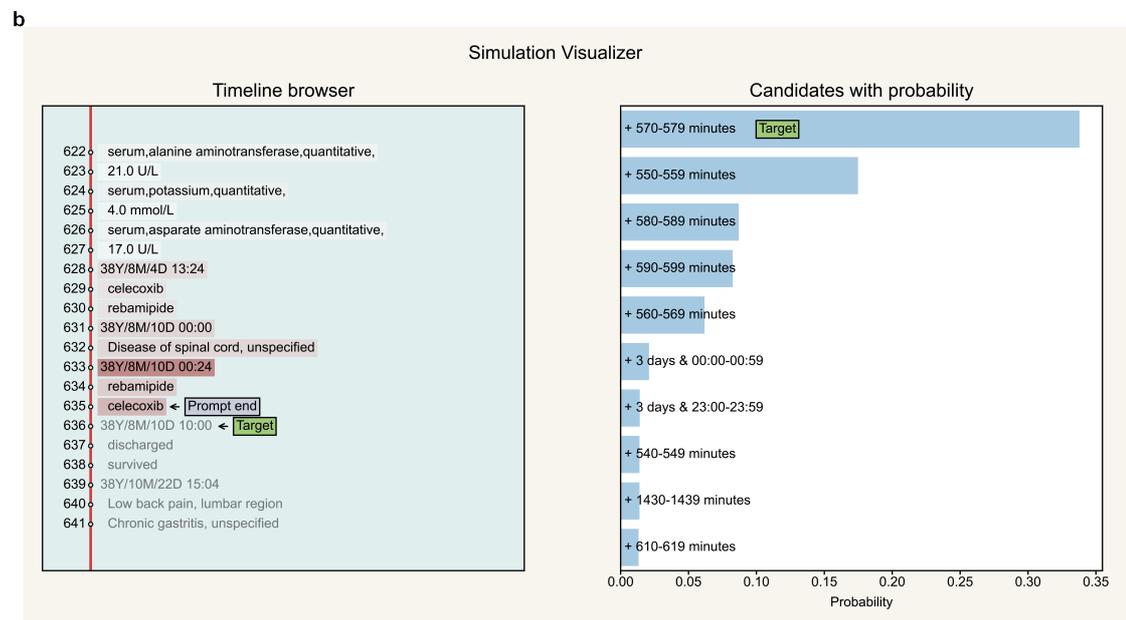

**Supplementary Fig. 2 | Prediction of numeric test results and time progressions.** This figure displays prediction steps made by the model. The left panels (timeline browser) show the clinical timeline for test patients: entries included in the prompt (model input) appear in dark text, whereas future entries are greyed out. Reddish highlights indicate attention weights. The right panels show predicted probability distributions over the model's vocabulary. **a**, Prediction of a numeric test result with strong dependency on other tests. In this example, the model predicts mean corpuscular haemoglobin (MCH). It assigns high probability to a narrow value range that includes the ground truth. Strong attention weights are placed on haemoglobin and red blood cell count, indicating their influence. This reflects the clinical formula MCH = (haemoglobin × 10)/RBC count, suggesting the model has implicitly learned this physiological relationship. **b**, Prediction of time progression. The model outputs a probability distribution over discretized time tokens. One token is selected, and actual time advancement is uniformly sampled within its defined range. In this example, the model favours time windows corresponding to the next morning (+570–599 minutes), when hospital discharge typically occurs. Attention weights indicate that the prediction was guided by recent diagnosis and prescription patterns.

| Record type | Patient age | Code / token | Entry name | Value / status |
|---|---|---|---|---|
| laboratory test | 3 days 07:00:00 | 3H0800000019---51 | whole blood,blood gas analysis,pH, | 7.464 no_unit |
| laboratory test | 3 days 07:00:00 | 3H0800000019---52 | whole blood,blood gas analysis,pCO2, | 28.5 mmHg |
| laboratory test | 3 days 07:00:00 | 3H0800000019---53 | whole blood,blood gas analysis,pO2, | 79.4 mmHg |
| laboratory test | 3 days 07:00:00 | 3H0800000019---54 | whole blood,blood gas analysis,actual bicarbonate, | 20.2 mmol/L |
| laboratory test | 3 days 07:00:00 | 3H0800000019---55 | whole blood,blood gas analysis,base excess, | -1.9 mmol/L |
| laboratory test | 3 days 07:00:00 | 3H0800000019---56 | whole blood,blood gas analysis,oxygen saturation, | 98.3 % |
| laboratory test | 3 days 07:00:00 | 3H0800000019---58 | whole blood,blood gas analysis,hemoglobin, | 14.6 g/dL |
| laboratory test | 3 days 07:00:00 | 3H0800000019---59 | whole blood,blood gas analysis,hematocrit, | 44.9 % |
| laboratory test | 3 days 07:00:00 | 3H0800000019---81 | whole blood,blood gas analysis,ionized sodium, | 141.0 mmol/L |
| laboratory test | 3 days 07:00:00 | 3H0800000019---82 | whole blood,blood gas analysis,ionized pottasium, | 4.4 mmol/L |
| laboratory test | 3 days 07:00:00 | 3H0800000019---83 | whole blood,blood gas analysis,ionized chlorine, | 110.0 mmol/L |
| laboratory test | 3 days 07:00:00 | 3H0800000019---84 | whole blood,blood gas analysis,ionized calcium, | 1.29 mmol/L |
| laboratory test | 3 days 07:00:00 | 3H0800000019---85 | whole blood,blood gas analysis,glucose, | 70.0 mg/dL |
| laboratory test | 3 days 07:00:00 | 3H0800000019---86 | whole blood,blood gas analysis,lactate, | 1.8 mmol/L |
| laboratory test | 3 days 10:04:00 | 2A9900000019---51 | whole blood,CBC,red blood cell count, | 329.0 ×10^4/μL |
| laboratory test | 3 days 10:04:00 | 2A9900000019---52 | whole blood,CBC,white blood cell count, | 4.6 ×10^3/μL |
| laboratory test | 3 days 10:04:00 | 2A9900000019---53 | whole blood,CBC,hemoglobin, | 13.8 g/dL |
| laboratory test | 3 days 10:04:00 | 2A9900000019---54 | whole blood,CBC,hematocrit, | 40.1 % |
| laboratory test | 3 days 10:04:00 | 2A9900000019---55 | whole blood,CBC,platelet count, | 22.1 ×10^4/μL |
| laboratory test | 3 days 10:04:00 | 2A9900000019---56 | whole blood,CBC,mean corpuscular volume, | 122.1 fL |
| laboratory test | 3 days 10:04:00 | 2A9900000019---57 | whole blood,CBC,mean corpuscular hemoglobin, | 42.0 pg |
| laboratory test | 3 days 10:04:00 | 2A9900000019---58 | whole blood,CBC,mean corpuscular hemoglobin concentration, | 34.4 % |
| laboratory test | 3 days 16:57:00 | 3H0800000019---51 | whole blood,blood gas analysis,pH, | 7.346 no_unit |
| laboratory test | 3 days 16:57:00 | 3H0800000019---52 | whole blood,blood gas analysis,pCO2, | 43.0 mmHg |
| laboratory test | 3 days 16:57:00 | 3H0800000019---53 | whole blood,blood gas analysis,pO2, | 36.4 mmHg |
| laboratory test | 3 days 16:57:00 | 3H0800000019---54 | whole blood,blood gas analysis,actual bicarbonate, | 22.8 mmol/L |
| laboratory test | 3 days 16:57:00 | 3H0800000019---55 | whole blood,blood gas analysis,base excess, | -2.5 mmol/L |
| laboratory test | 3 days 16:57:00 | 3H0800000019---56 | whole blood,blood gas analysis,oxygen saturation, | 76.6 % |
| laboratory test | 3 days 16:57:00 | 3H0800000019---58 | whole blood,blood gas analysis,hemoglobin, | 17.0 g/dL |
| laboratory test | 3 days 16:57:00 | 3H0800000019---59 | whole blood,blood gas analysis,hematocrit, | 52.0 % |
| laboratory test | 3 days 16:57:00 | 3H0800000019---81 | whole blood,blood gas analysis,ionized sodium, | 137.0 mmol/L |
| laboratory test | 3 days 16:57:00 | 3H0800000019---82 | whole blood,blood gas analysis,ionized pottasium, | 4.4 mmol/L |
| laboratory test | 3 days 16:57:00 | 3H0800000019---83 | whole blood,blood gas analysis,ionized chlorine, | 110.0 mmol/L |
| laboratory test | 3 days 16:57:00 | 3H0800000019---84 | whole blood,blood gas analysis,ionized calcium, | 1.27 mmol/L |
| laboratory test | 3 days 16:57:00 | 3H0800000019---85 | whole blood,blood gas analysis,glucose, | 85.0 mg/dL |
| laboratory test | 3 days 16:57:00 | 3H0800000019---86 | whole blood,blood gas analysis,lactate, | 2.0 mmol/L |
| drug order | 4 days 13:48:00 | B02BA | Vitamin K | |
| diagnosis | 4 days 16:21:00 | P616 | Other transient neonatal disorders of coagulation | |
| laboratory test | 4 days 17:30:00 | 3H0800000019---51 | whole blood,blood gas analysis,pH, | 7.346 no_unit |
| laboratory test | 4 days 17:30:00 | 3H0800000019---52 | whole blood,blood gas analysis,pCO2, | 44.0 mmHg |
| laboratory test | 4 days 17:30:00 | 3H0800000019---53 | whole blood,blood gas analysis,pO2, | 43.1 mmHg |
| laboratory test | 4 days 17:30:00 | 3H0800000019---54 | whole blood,blood gas analysis,actual bicarbonate, | 23.4 mmol/L |
| laboratory test | 4 days 17:30:00 | 3H0800000019---55 | whole blood,blood gas analysis,base excess, | -1.9 mmol/L |
| laboratory test | 4 days 17:30:00 | 3H0800000019---56 | whole blood,blood gas analysis,oxygen saturation, | 83.2 % |
| laboratory test | 4 days 17:30:00 | 3H0800000019---58 | whole blood,blood gas analysis,hemoglobin, | 15.0 g/dL |
| laboratory test | 4 days 17:30:00 | 3H0800000019---59 | whole blood,blood gas analysis,hematocrit, | 46.0 % |
| laboratory test | 4 days 17:30:00 | 3H0800000019---81 | whole blood,blood gas analysis,ionized sodium, | 139.0 mmol/L |
| laboratory test | 4 days 17:30:00 | 3H0800000019---82 | whole blood,blood gas analysis,ionized pottasium, | 4.4 mmol/L |
| laboratory test | 4 days 17:30:00 | 3H0800000019---83 | whole blood,blood gas analysis,ionized chlorine, | 107.0 mmol/L |
| laboratory test | 4 days 17:30:00 | 3H0800000019---84 | whole blood,blood gas analysis,ionized calcium, | 1.36 mmol/L |
| laboratory test | 4 days 17:30:00 | 3H0800000019---85 | whole blood,blood gas analysis,glucose, | 81.0 mg/dL |
| laboratory test | 4 days 17:30:00 | 3H0800000019---86 | whole blood,blood gas analysis,lactate, | 1.6 mmol/L |
| drug order | 4 days 20:50:00 | B02BA | Vitamin K | |
| discharge event | 5 days 10:42:00 | [DSC] | discharged | survived |

**Supplementary Fig. 3| The pretrained model generated realistic clinical timelines.** This example shows a timeline generated by the pretrained model using a prompt from a newborn patient on their second day of hospitalization. The model continued generation until hospital discharge. Timestamps are shown at minute-level resolution. Laboratory test results are reported with appropriate units and decimal precision. The model also reproduces physiologically consistent relationships between test results. For example, the generated mean corpuscular volume (MCV) is 122.0 fL. Theoretical MCV can be calculated as haematocrit (Ht) divided by red blood cell count (RBC), multiplied by 1,000: Ht ÷ RBC × 1,000 = 40.1 ÷ 329.0 × 1,000 = 121.9. Similarly, mean corpuscular haemoglobin (MCH) and mean corpuscular haemoglobin concentration (MCHC) are derived from haemoglobin (Hb): MCH = Hb ÷ RBC × 1,000 = 41.9; MCHC = Hb ÷ Ht × 100 = 34.4 These theoretical values closely match the model's outputs (MCH = 42.0, MCHC = 34.4), demonstrating that the model preserves intervariable consistency in laboratory test results.

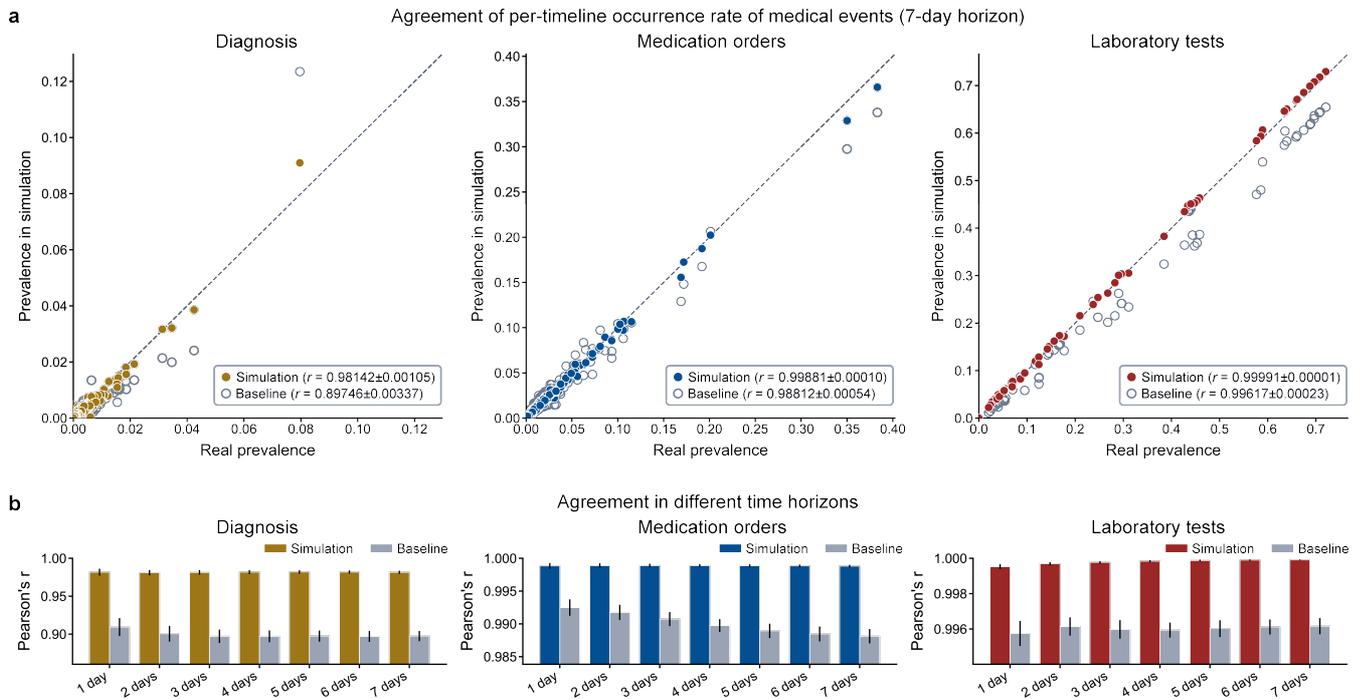

**Supplementary Fig. 4 | The simulator accurately reproduced medical event rates at per-timeline level.** The model demonstrated high fidelity in the overall event frequency (Fig. 3). However, this metric can be disproportionally represented by long timelines that contain a large volume of medical events. To assess the model's performance at the individual timeline level, we evaluated the per-timeline event rates, and this figure shows the results. This prevalence metric was defined as the proportion of individual timelines in which a given event appeared at least once. **a**, Agreement of timeline-level occurrence rate of medical events in the 7-day time horizon. **b**, Agreement of timeline-level occurrence rates in real-world versus simulation data across different time horizons.

8
9

## Supplementary Table 2 | Laboratory tests included in the correlation analysis

| Number | Test name | Specimen | Essential general IVD | Code |
|---|---|---|---|---|
| 1 | CBC (red blood cell count) | whole blood | ✓ | 2A9900000019---51 |
| 2 | CBC (white blood cell count) | whole blood | ✓ | 2A9900000019---52 |
| 3 | CBC (haemoglobin) | whole blood | ✓ | 2A9900000019---53 |
| 4 | CBC (haematocrit) | whole blood | ✓ | 2A9900000019---54 |
| 5 | CBC (mean corpuscular volume) | whole blood | ✓ | 2A9900000019---56 |
| 6 | CBC (mean corpuscular haemoglobin) | whole blood | ✓ | 2A9900000019---57 |
| 7 | CBC (mean corpuscular haemoglobin concentration) | whole blood | ✓ | 2A9900000019---58 |
| 8 | CBC (platelet count) | whole blood | ✓ | 2A9900000019---55 |
| 9 | Creatinine (quantitative) | serum | ✓ | 3C0150000023---01 |
| 10 | Alanine aminotransferase (quantitative) | serum | ✓ | 3B0450000023---01 |
| 11 | Estimated GFR based on creatinine (quantitative) | serum | ✓ | 8A0650000023---01 |
| 12 | Aspartate aminotransferase (quantitative) | serum | ✓ | 3B0350000023---01 |
| 13 | Blood urea nitrogen (quantitative) | serum | ✓ | 3C0250000023---01 |
| 14 | CBC (RBC distribution width, coefficient of variation) | whole blood | ✓ | 2A9900000019---60 |
| 15 | CBC (platelet crit) | whole blood | ✓ | 2A9900000019---63 |
| 16 | CBC (mean platelet volume) | whole blood | ✓ | 2A9900000019---61 |
| 17 | CBC (platelet distribution width) | whole blood | ✓ | 2A9900000019---62 |
| 18 | Gamma glutamyl transpeptidase (quantitative) | serum | ✓ | 3B0900000023---01 |
| 19 | Potassium (quantitative) | serum | ✓ | 3H0150000023---01 |
| 20 | Albumin (quantitative) | serum | ✓ | 3A0150000023---01 |
| 21 | C-reactive protein (quantitative) | serum | ✓ | 5C0700000023---01 |
| 22 | Sodium (quantitative) | serum | ✓ | 3H0100000023---01 |
| 23 | Lactate dehydrogenase (quantitative) | serum | ✓ | 3B0500000023---01 |
| 24 | Total bilirubin (quantitative) | serum | ✓ | 3J0100000023---01 |
| 25 | Blood urea nitrogen (creatinine-adjusted) | serum | ✓ | 3C0250000023---28 |
| 26 | Haemogram (basophil) | whole blood | ✓ | 2A1600000019---55 |
| 27 | Haemogram (eosinophil) | whole blood | ✓ | 2A1600000019---54 |
| 28 | Haemogram (monocyte) | whole blood | ✓ | 2A1600000019---56 |
| 29 | Haemogram (lymphocyte) | whole blood | ✓ | 2A1600000019---57 |
| 30 | Alkaline phosphatase (quantitative) | serum | ✓ | 3B0700000023---01 |
| 31 | Haemogram (neutrophil) | whole blood | ✓ | 2A1600000019---51 |
| 32 | Chloride (quantitative) | serum | ✓ | 3H0200000023---01 |
| 33 | Uric acid (quantitative) | serum | ✓ | 3C0200000023---01 |
| 34 | Total proteins (quantitative) | serum | ✓ | 3A0100000023---01 |
| 35 | Creatine kinase (quantitative) | serum |  | 3B0100000023---01 |
| 36 | Albumin/globulin ratio (composition ratio) | serum |  | 3A0160000023---02 |
| 37 | Calcium (quantitative) | serum | ✓ | 3H0300000023---01 |
| 38 | Prothrombin time (international normalized ratio) | plasma | ✓ | 2B0300000022---57 |
| 39 | Prothrombin time (PT activity, %) | plasma | ✓ | 2B0300000022---53 |
| 40 | Prothrombin time (coagulation time sample, sec) | plasma | ✓ | 2B0300000022---51 |
| 41 | Prothrombin time (PT ratio) | plasma | ✓ | 2B0300000022---55 |
| 42 | Glucose(quantitative) | plasma | ✓ | 3D0100000022---01 |
| 43 | Adjusted calcium (quantitative) | serum | ✓ | 3H0380000023---01 |
| 44 | Creatinine (quantitative) | urine | ✓ | 3C0150000001---01 |
| 45 | LDL-cholesterol (quantitative) | serum | ✓ | 3F0770000023---01 |
| 46 | Urinalysis qualitative/semiquantitative (pH) | urine | ✓ | 1A9900000001---52 |
| 47 | Triglyceride (quantitative) | serum | ✓ | 3F0150000023---01 |
| 48 | Activated partial thromboplastin time (coagulation time sample, sec) | plasma | ✓ | 2B0200000022---51 |
| 49 | Cholesterol total (quantitative) | serum | ✓ | 3F0500000023---01 |
| 50 | Albumin (quantitative) | urine | ✓ | 3A0150000001---01 |
| 51 | Amylase (quantitative) | serum | ✓ | 3B1600000023---01 |
| 52 | Haemoglobin A1c (composition ratio) | whole blood |  | 3D0460000019---02 |
| 53 | HDL-cholesterol (quantitative) | serum | ✓ | 3F0700000023---01 |
| 54 | Inorganic phosphate (quantitative) | serum | ✓ | 3H0400000023---01 |
| 55 | Conjugated bilirubin (quantitative) | serum | ✓ | 3J0150000023---01 |
| 56 | CBC (immature platelet fraction) | whole blood | ✓ | 2A9900000019---66 |
| 57 | Fibrinogen (quantitative) | plasma | ✓ | 2B1000000022---01 |
| 58 | D-dimer (quantitative) | plasma | ✓ | 2B1400000022---01 |
| 59 | Urine specific gravity (quantitative) | urine | ✓ | 1A0300000001---01 |
| 60 | Non-HDL-cholesterol (quantitative) | serum | ✓ | 3F0690000023---01 |
| 61 | Carcinoembryonic antigen (quantitative) | serum |  | 5D0100000023---01 |
| 62 | Haemogram (fragmented cell, %) | whole blood | ✓ | 2A1600000019---06 |
| 63 | Brain natriuretic polypeptide (quantitative) | plasma |  | 4Z2710000022---01 |
| 64 | CA19-9 (quantitative) | serum |  | 5D1300000023---01 |

| 65 | Thyroid-stimulating hormone (quantitative) | serum | | 4A0550000023---01 |
| 66 | Free thyroxine (quantitative) | serum | | 4B0350000023---01 |
| 67 | Magnesium (quantitative) | serum | ✓ | 3H0250000023---01 |
| 68 | Haemogram (stab form neutrophil) | whole blood | ✓ | 2A1600000019---52 |
| 69 | Haemogram (segmented neutrophil) | whole blood | ✓ | 2A1600000019---53 |
| 70 | Haemogram (myelocyte) | whole blood | ✓ | 2A1600000019---59 |
| 71 | Haemogram (myeloblast) | whole blood | ✓ | 2A1600000019---62 |
| 72 | Haemogram (promyelocyte) | whole blood | ✓ | 2A1600000019---61 |
| 73 | Haemogram (metamyelocyte) | whole blood | ✓ | 2A1600000019---60 |
| 74 | Haemogram (atypical lymphocyte) | whole blood | ✓ | 2A1600000019---58 |
| 75 | Haemogram (others) | whole blood | ✓ | 2A1600000019---77 |
| 76 | Blood gas analysis (pH) | whole blood | ✓ | 3H0800000019---51 |
| 77 | Blood gas analysis (actual bicarbonate) | whole blood | ✓ | 3H0800000019---54 |
| 78 | Blood gas analysis (base excess) | whole blood | ✓ | 3H0800000019---55 |
| 79 | Blood gas analysis ($pCO_2$) | whole blood | ✓ | 3H0800000019---52 |
| 80 | Blood gas analysis (haemoglobin) | whole blood | ✓ | 3H0800000019---58 |
| 81 | Blood gas analysis (haematocrit) | whole blood | ✓ | 3H0800000019---59 |
| 82 | Blood gas analysis ($pO_2$) | whole blood | ✓ | 3H0800000019---53 |
| 83 | Blood gas analysis (oxygen saturation) | whole blood | ✓ | 3H0800000019---56 |
| 84 | Blood gas analysis (lactate) | whole blood | ✓ | 3H0800000019---86 |
| 85 | Blood gas analysis (ionized sodium) | whole blood | ✓ | 3H0800000019---81 |

This list shows the laboratory tests used for intertest correlation analyses. Tests are ordered by frequency, accounting for 90% of the total test volume in the training data. Items compatible with the general IVDs listed in the 4th WHO Model List of Essential In Vitro Diagnostics are marked with check marks. CBC, complete blood count; GFR, glomerular filtration rate; HDL, high-density lipoprotein; IVD, in vitro diagnostic; LDL, low-density lipoprotein; PT, prothrombin time; RBC, red blood cell; WHO, World Health Organization.

10  The list of codes sufficiently covers routinely ordered general in vitro diagnostics (IVDs). The WHO
11  publishes the *Model List of Essential In Vitro Diagnostics (EDL)*, which includes essential clinical chemistry
12  and haematology IVDs recommended for use in clinical laboratories. Our list includes nearly all these tests,
13  with only eight exceptions. These exceptions are blood crossmatching (used for transfusion preparation), blood
14  culture (for blood stream infection), Coombs test (for haemolytic conditions), human chorionic gonadotropin
15  (for pregnancy assessment), iron studies (for anaemia evaluation), procalcitonin (for infection or sepsis
16  assessment), sickle cell testing (for haemoglobinopathies), and troponin (for cardiac injury). These tests are
17  ordered under specific clinical indications rather than as part of routine testing. As such, our list sufficiently
18  represents routinely ordered general IVDs as defined in the WHO EDL.
19
20

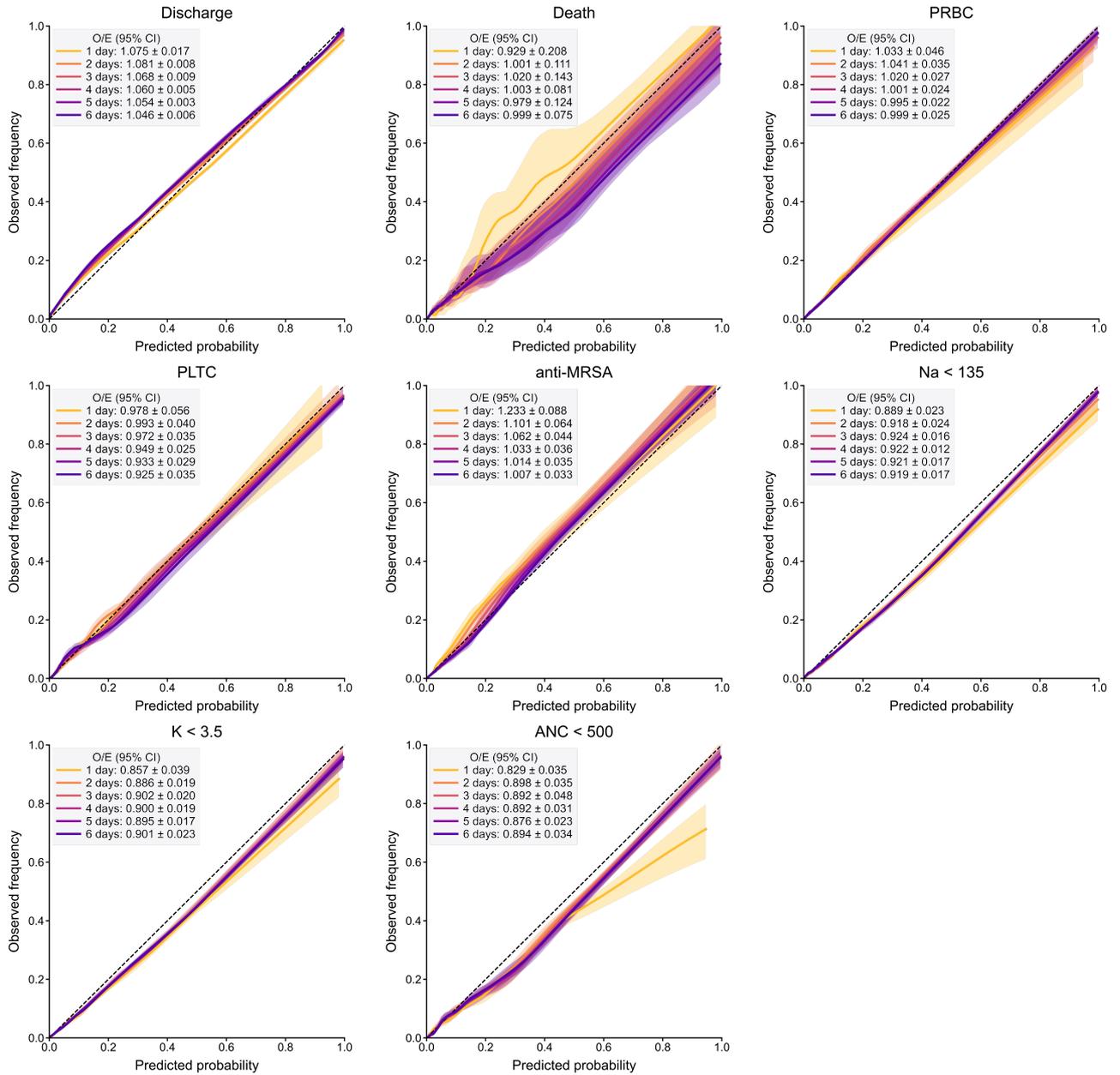

**Supplementary Fig. 5 | The simulator reproduced clinical event probabilities across various outcomes and time horizons.** The calibration curves for the selected clinical events are displayed. Shaded areas indicate 95% confidence intervals. The model showed stable calibrations across different clinical events and time horizons (1–6 days). ANC, absolute neutrophil count; K, potassium; Na, sodium; MRSA, methicillin-resistant *Staphylococcus aureus*; O/E ratio, observed-to-expected ratio, PRBC, packed red blood cells; PLTC, platelet concentrate.

21
22

# Supplementary methods
## Data cleaning

We mapped the diagnosis codes to the 10th Revision of the International Classification of Diseases (ICD-10). We used the standardized Japanese diagnosis code system *Disease Management ID*, which is widely employed in HL7 messages and insurance claims in Japan. This system is designed for compatibility with the ICD-10 and is maintained by the Medical Information System Development Center (MEDIS), Japan's official coding authority. Mapping was performed by merging diagnosis records with the latest conversion table published by MEDIS.

Medication codes were mapped to the Anatomical Therapeutic Chemical (ATC) Classification System. We used standardized Japanese medication codes (YJ and HOT9 codes). The initial mapping used conversion tables provided by the Japan Pharmaceutical Information Center (JAPIC). For records that remained unmapped after this process, we developed an auxiliary mapping table. First, we extracted unique product names from these records. ATC codes were then manually assigned by referencing the JAPIC's official tables and the ATC index from the World Health Organization. This supplemental mapping table was used to complete the alignment of all medication records with ATC codes.

We standardized laboratory test units to ensure internal consistency. For tests reported in multiple units, we applied unit conversion when appropriate. For example, white blood cell counts reported in /μL were divided by 1,000 and converted to ×1,000/μL. Conversions were based on a predefined table derived from a laboratory unit standardization list published by MEDIS. If a recommended unit was provided in this list, it was adopted as the standard. When no standard unit was specified, we used the most frequently observed unit for each test code. If a unit conversion required operations that were more complex than arithmetic scaling (e.g., nonlinear transformations), we retained the original units to avoid introducing conversion errors.

We also standardized nonnumeric laboratory results. All values were first normalized using Unicode Normalization Form Compatibility Composition (NFKC). We then mapped values indicating positivity (e.g., '+', 'positive', '(+)') to a unified label '(+)'; values indicating negativity (e.g., '−', 'negative', '(−)') to '(−)'; and borderline or ambiguous results (e.g., '+/-', 'equivocal') to '(+/−)'. In addition, entries representing missing or invalid results (e.g., 'NA', 'reported elsewhere', 'invalid specimen') were mapped to null. Records containing such null values were excluded from downstream analyses. Original values reported in Japanese were translated into English to ensure consistency (e.g., 'クラスIII' was translated to 'class III'). The remaining nonnumeric results not covered by these categories were retained in their original form.

## 55 List of special tokens and subtokens

### Supplementary Table 1 | Predefined special tokens.

|  | Description | Vocabulary |
|---|---|---|
| **Tokens** | | |
| [PAD] | Padding | |
| [F] | Female sex | ✓ |
| [M] | Male sex | ✓ |
| [A] | Ambiguous sex | ✓ |
| [U] | Unknown sex | ✓ |
| [N] | Sex not applicable | ✓ |
| [O] | Other sex | ✓ |
| [ADM] | Admission | ✓ |
| [DSC] | Discharge | ✓ |
| [DSC_ALV] | Discharge outcome of survival | ✓ |
| [DSC_EXP] | Discharge outcome of death | ✓ |
| [EOT] | End of timeline | ✓ |
| [POS] | Positive laboratory test result | ✓ |
| [NEG] | Negative laboratory test result | ✓ |
| [N/P] | Equivocal laboratory test result | ✓ |
| **Sub-tokens** | | |
| [DX] | Prepended to diagnoses | |
| [MED] | Prepended to medications | |
| [LAB] | Prepended to laboratory tests | |
| [PRV] | Appended to provisional diagnoses | |

Tokens with check marks are distinct vocabulary entries of the model.

## 56

57 **Discretization of time progression**

58

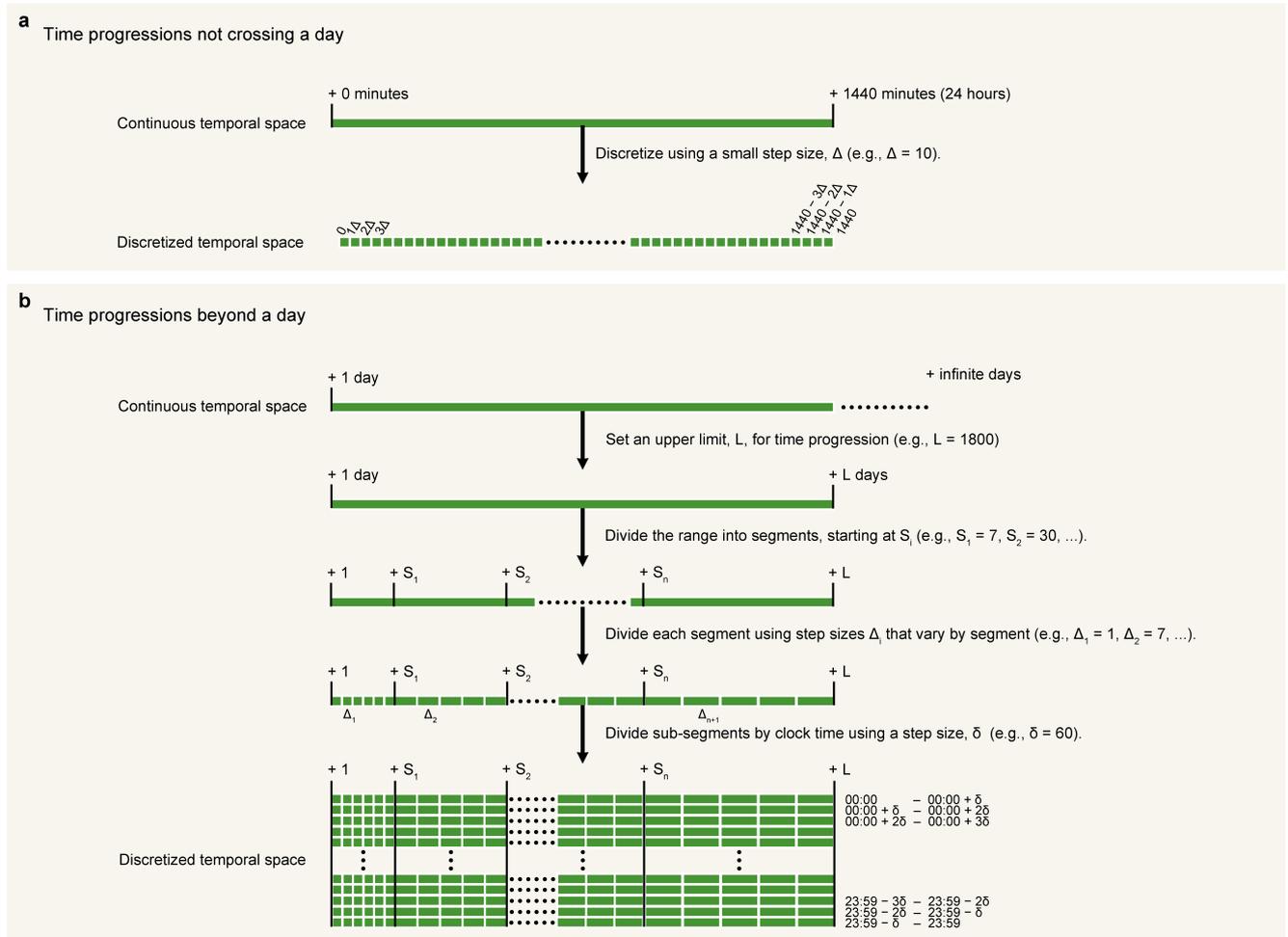

**Supplementary Fig. 6. Discretization schema of temporal progression. a.** Time intervals within 24 hours were discretized into uniform bins. We used a bin width of $\Delta = 10$ days. **b.** For time intervals beyond 24 hours, we applied a hierarchical discretization. First, an upper limit was defined to bound the temporal range ($L = 1800$ days). This range was divided into variable-length segments. In this study, we used $S_1 = 30$, $S_2 = 180$, $S_3 = 360$ days. Each segment was further subdivided using finer resolutions, using $\Delta_1 = 1$, $\Delta_2 = 10$, $\Delta_3 = 30$, $\Delta_4 = 180$ (days). Finally, each subsegment was partitioned into small uniform bins based on clock time. We used $\delta = 60$ (minutes). All bins are right-exclusive (i.e., '[a, b)' form).

59